\documentclass[acmsmall]{acmart}

\usepackage{color,soul}
\usepackage{soulpos}
\usepackage{soul}
\usepackage{bm,nicefrac}
\usepackage[strings]{underscore}
\usepackage{algorithm}
\usepackage{algpseudocode}
\usepackage{graphicx}
\usepackage{hyperref}
\usepackage{tipa}
\usepackage[T1]{fontenc}
\usepackage{libertine}
\usepackage{newtxmath}

\AtBeginDocument{
  }

\setcopyright{acmlicensed}
\copyrightyear{2025}
\acmYear{2025}
\acmDOI{XXXXXXX.XXXXXXX}

\acmJournal{JACM}
\acmVolume{37}
\acmNumber{4}
\acmArticle{111}
\acmMonth{8}

\begin{document}

\title{JambaTalk: Speech-driven 3D Talking Head Generation based on a Hybrid Transformer-Mamba Model}

\author{Farzaneh Jafari}
\orcid{0009-0007-1901-2860}
\affiliation{
  \institution{University of Alberta, Multimedia Research Center (MRC)}
  \city{Edmonton, AB}
  \state{Alberta}
  \country{Canada}}
\email{farzane1@ualberta.ca}

\author{Stefano Berretti}
\affiliation{
  \institution{University of Florence, Media Integration and Communication Center (MICC)}
  \city{Florence}
  \country{Italy}}
\email{stefano.berretti@unifi.it}

\author{Anup Basu}
\affiliation{
 \institution{University of Alberta, Multimedia Research Center (MRC)}
 \city{Edmonton, AB}
 \state{Alberta}
 \country{Canada}}
\email{basu@ualberta.ca}

\begin{abstract}
    In recent years, the talking head generation has become a focal point for researchers. Considerable effort is being made to refine lip-sync motion, capture expressive facial expressions, generate natural head poses, and achieve high-quality video. However, no single model has yet achieved equivalence across all quantitative and qualitative metrics. We introduce JambaTalk, a hybrid Transformer-Mamba model, to animate a 3D face. Mamba, a pioneering Structured State Space Model (SSM) architecture, was developed to overcome the limitations of conventional Transformer architectures, particularly in handling long sequences. This challenge has constrained traditional models. Jamba combines the advantages of both the Transformer and Mamba approaches, offering a comprehensive solution. Based on the foundational Jamba block, we present JambaTalk to enhance motion variety and lip sync through multimodal integration. Extensive experiments reveal that our method achieves performance comparable or superior to state-of-the-art models. Supplementary video and code are available at: \href{https://farzanehjafari1987.github.io/JambaTalk.github.io/}{https://JambaTalk.github.io/}. 
\end{abstract}

\begin{CCSXML}
<ccs2012>
   <concept>
       <concept_id>10010147.10010371.10010352</concept_id>
       <concept_desc>Computing methodologies~Animation</concept_desc> 
       <concept_significance>500</concept_significance>
       </concept>
   <concept>
       <concept_id>10010147.10010178</concept_id>
       <concept_desc>Computing methodologies~Artificial intelligence</concept_desc>
       <concept_significance>500</concept_significance>
       </concept>
   <concept>
       <concept_id>10010147.10010371.10010382</concept_id>
       <concept_desc>Computing methodologies~Image manipulation</concept_desc>
       <concept_significance>500</concept_significance>
       </concept>
 </ccs2012>
\end{CCSXML}

\ccsdesc[500]{Computing methodologies~Animation}
\ccsdesc[500]{Computing methodologies~Artificial intelligence}
\ccsdesc[500]{Computing methodologies~Image manipulation}

\keywords{Facial Animation Synthesis, Virtual Humans, 3D Mesh Animation, Blendshape Animation, Selective State Space Models (S6), Mixture of Experts, Transformers, Mamba, and Low-Rank Learned Rotary Positional Embedding.}

\maketitle

\section{Introduction}
Talking-head generation has gained increasing attention for applications in virtual avatars, gaming, online education, and communication. Recent advances have improved lip-sync accuracy, natural facial motion, and high-quality 3D avatar synthesis. However, many existing approaches rely on Transformer-based architectures~\cite{fan2022faceformer, xing2023codetalker, Peng2023Selftalk}, which face limitations when modeling long sequences due to high computational complexity and memory requirements. This challenge restricts scalability and efficiency, particularly when looking for real-time applications or large-scale datasets.

Many existing methods focus primarily on lip-sync accuracy from audio input~\cite{Liu2023Moda, Zhong2023Identity-preserving, Xu2024Vase-1, Wei2024Aniportrait, Xu2024Hallo, Ye2023Geneface++}. Most audio-driven talking face generation approaches adopt a two-stage pipeline. In the first stage, an intermediate representation, such as 2D landmarks~\cite{Tang2022Memories, Shen2023Difftalk, Du2023Dae-talker, Wang2023Seeing, Guan2023Stylesync} or 3D face model blend-shape coefficients~\cite{Ko2024Talk3D, Su2023DualTalker, Nocentini2023Learning}, is predicted from the audio input. In the second stage, a renderer synthesizes the video portrait based on this representation. Audio-driven talking face generation aims to produce high-quality, lip-synchronized face videos from a given audio clip and target face image. However, this task remains challenging given the inherent differences between audio and visual modalities. The generation of a one-shot 3D talking portrait further increases the complexity for attempting to create a 3D avatar from a single unseen image and animating it using a reference video or audio~\cite{Ye2024Real3d-portrait}. Moreover, face-to-face spoken dialogue models capture audio-visual speech input from the user and generate audio-visual responses, serving as a first step toward avatar-based chatbots that operate without intermediate text~\cite{Sun2022Masked, Goyal2023Emotionally, Tan2024Style2talker, Tan2024Say, Tan2024EDTalk}. Another open challenge lies in generalizing talking-face models to multiple languages and diverse text inputs~\cite{anderson2013an, Song2022Talking}.

State-Space Models (SSMs) have recently demonstrated significant potential for addressing long-range dependency issues~\cite{Gu2023Mamba}, but they still fall short of the performance achieved by comparably sized Transformer-based models. To overcome these limitations, we introduce JambaTalk, built upon Jamba, a publicly available large language model that employs a hybrid architecture combining Transformer layers with Mamba layers, along with a Mixture-of-Experts (MoE) mechanism. This combination leverages the strengths of attention (for global context modeling) and SSMs (for efficient long-sequence processing), resulting in improved performance, higher throughput, and reduced memory footprint compared to traditional Transformer-only models. The key novelty of Jamba lies in its hybrid design, which balances memory efficiency, long-context handling (up to 256K tokens), and scalability. MoE layers are incorporated into the MLP components, expanding the total model capacity without increasing the computational cost. This design enables Jamba to handle extremely long sequences efficiently, where attention-only models often struggle~\cite{Lieber2024Jamba}. Our implementation runs efficiently on a single 24GB GPU, demonstrating high throughput, low memory usage, and state-of-the-art performance in long-context evaluations. The main contributions of our work are summarized below:
\begin{itemize}
    \item We introduce JambaTalk, an innovative framework for speech-driven 3D talking head generation that cooperates with multiple Mamba, MoE\_Mamba, and Transformer layers to improve the performance of talking face generation;
    \item We introduce Low-Rank Learned Rotary Positional Embedding (LRL-RoPE), a learnable and low-memory extension of RoPE, and take advantage of the potential of the Group Query Attention (GQA) algorithm to enhance the performance of the Transformer layer; 
    \item Comprehensive experiments and analyses on the Vocaset and \(BIWI_6\) datasets demonstrate the effectiveness of the proposed model.
\end{itemize}

\section{Related Work}

\subsection{Speech-Driven 3D Facial Animation}
Speech-driven 3D facial animation has seen substantial progress in recent years, with methods focusing on generating expressive and synchronized facial movements from audio~\cite{chen2023hyperlips}. This research area spans traditional neural architectures, diffusion-based models, codebook-based synthesis, and controllable mesh deformation.

VOCA~\cite{Cudeiro2019Capture} introduced a neural network trained on a unique 4D face dataset, enabling identity-independent facial animation from speech. FaceFormer~\cite{fan2022faceformer} applied a Transformer-based auto-regressive architecture to capture long-term audio context, enhanced by pre-trained speech representations and bias-aware attention mechanisms. 
MeshTalk~\cite{Richard2021Meshtalk} introduced a categorical latent space that separates audio-driven and audio-independent features, enabling high-quality full-face animations including blinks and eyebrow motion. Xing et al.~\cite{xing2023codetalker} reformulated facial animation as a code query problem over a learned motion codebook, improving the vividness and accuracy of generated expressions.

Diffused Heads~\cite{Stypułkowski2024Diffused} demonstrated the effectiveness of diffusion-based generative models in producing realistic talking heads using only audio and a single image. FaceDiffuser~\cite{Stan2023Facediffuser} extended this by supporting both vertex and blendshape datasets, improving compatibility with real-world animation pipelines. 
VividTalk~\cite{Sun2023Vividtalk} proposed a two-stage mesh-to-video pipeline integrating non-rigid expression and rigid motion learning, with a head pose codebook and dual-branch motion decoding. Zou et al.~\cite{Zou20234D} presented a landmark-guided DDPM framework for generating 3D mesh sequences, enabling conditional generation from text, geometry, or partial sequences.

Rai et al.~\cite{Rai2024Towards} incorporated 2D generative models into a controllable 3D face framework for semantic manipulation. Otberdout et al.~\cite{Otberdout2020Dynamic, Otberdout2022Sparse, Otberdout2023Generating} introduced a motion transition framework combining temporal GANs (Motion3DGAN) and a decoder (S2D-Dec) to generate dense point clouds from sparse 3D landmark motions between emotional states.

While early work focused primarily on lip-sync, recent efforts address expressive control and emotional realism in talking head generation. These methods incorporate emotion recognition, disentanglement, and continuous expression control. Speech Emotion Recognition (SER) plays a crucial role in realistic animation. Traditional SER approaches include handcrafted audio features~\cite{liang2022expressive, Schuller2003Hidden, Kwon2003Emotion}, while recent deep learning methods use end-to-end training for emotional feature extraction~\cite{gan2023efficient, gururani2023space, Ji2022Emma, saunders2023read, tan2024flowvqtalker, tian2024emo, xu2024facechain-imagineid, Papantoniou2022Neural, Huang2014Speech, Jain2020Speech, Schuller2018Speech, Hossain2019Emotion, LeCun1998Gradient, Mekruksavanich2020Negative}. 
EmoVOCA~\cite{Nocentini2024EmoVOCA} and DEITalk~\cite{Shen2024DEITalk} integrated SER into facial animation pipelines. 
EmoTalk~\cite{Peng2023Emotalk} introduced an Emotion Disentangling Encoder (EDE) and a fusion decoder to separate emotional and content cues for enhanced expressivity. 

EmoStyle~\cite{Azari2024EmoStyle} leveraged valence-arousal coordinates and a pre-trained StyleGAN2 generator to manipulate facial expressions across a wide emotional spectrum, supporting one-shot processing of real portraits. Sung-Bin et al.~\cite{Sung-Bin2024LaughTalk} addressed non-verbal expression by introducing LaughTalk, a model capable of generating both articulated speech and natural laughter. 
EMOTE~\cite{Daněček2023Emotional} proposed a framework for emotion-aware 3D avatars by separating high-frequency lip-sync deformation from low-frequency emotional expressions using decoupled loss functions.

\subsection{Selective State Space Models}
Transformer models are widely used in sequence modeling but struggle with high computational and memory demands, especially for long contexts. To address this, Selective State Space Models (SSMs) have been introduced.

Mamba~\cite{Gu2023Mamba} introduces a hardware-efficient, attention-free sequence model that adapts its internal parameters dynamically based on the input token stream. It overcomes conventional SSM limitations with a parallelized convolution-based architecture and avoids reliance on attention or MLP blocks. Mamba offers advantages in terms of memory efficiency and recurrent processing, while maintaining long-range dependency modeling. 

Xu et al.~\cite{Xu2024Mambatalk} extended Mamba into MambaTalk, targeting speech-driven full-body gesture synthesis. It employs discrete motion priors and multimodal integration to handle the limitations of diffusion-based and attention-based methods, producing natural and diverse body motions with efficient computation. These insights form a foundation for applying SSMs to expressive facial animation, as explored in our proposed JambaTalk framework.

\section{Proposed Method}
Our primary goal is to generate sequential 3D facial animations from raw audio input and prior facial movement sequences, resulting in realistic and synchronized animations. Suppose we are given a sequence of ground-truth 3D facial motions \(P_t=(p_1,...,p_T)\), where \(T\) denotes the number of visual frames sampled at a specific frame-per-second (FPS) rate for the dataset, alongside the corresponding raw audio signals \(\chi\). The purpose of the model is to generate synthesized facial movements \(\hat{P}_T\) that closely approximate \(P_T\), using the raw audio inputs \(\chi\)~\cite{fan2022faceformer, xing2023codetalker}. In this architecture, the process begins with audio encoders, which convert the raw audio signals into sequential speech representations \(A_T=(a_1,...,a_T)\), where \(T\) represents the frame length of the speech representation. This speech encoding captures the temporal and phonetic dynamics required for facial movement synthesis. The motion decoder predicts facial movements \(\hat{P}_T=(\hat{p}_1,...,\hat{p}_T)\) based on the encoded speech sequences \(A_T\). Generally, we have:
\begin{equation}
\hat{p}_t=JambaTalk(a_t; \eta), 
\end{equation}

\noindent
where \(\eta\) signifies the model parameters, \(t\) is the current time-step in the sequence, and \(\hat{p}_t \in \hat{P}_T\). 
Following~\cite{fan2022faceformer}, we utilize the pre-trained Wav2Vec~2.0 model~\cite{Baevski2020wav2vec} with fixed weights. The encoder includes an audio feature extractor and a multi-layer Transformer encoder. The audio feature extractor, consisting of several temporal convolutional layers (TCN), converts the raw waveform input into feature vectors with frequency \(f_a\). A linear interpolation layer is added for resampling the audio components using the captured frequency of facial motion \(f_m\) with the output length \(kT\), where \(k = \frac{f_m}{f_a}\) (e.g., if \(f_a = 50\) and \(f_m = 25\), then \(k = 0.5\), meaning the audio is sampled twice as fast and should be downsampled to match the facial motion rate). These components are combined with multi-head self-attention and feed-forward layers to transform the audio feature vectors into contextualized speech representations, in which the outputs of the TCN are discretized into a finite set of speech units by a quantization module. The linear projection layer outputs can be demonstrated as \(A_{kT} = (a_1, ..., a_{kT})\). Ultimately, the predicted facial motion \(\hat{p}_t\) is derived by mapping the \(d\)-dimensional hidden state back into the \(V\)-dimensional 3D vertex space using a motion decoder. The motion decoder utilizes multiple layers of masked self-attention and feed-forward neural networks to transform audio features into 3D mesh deformations. In this process, masked self-attention assigns weights to input embeddings based on their relevance, and the feed-forward network refines these context vectors to produce the final outputs. After generating the complete 3D facial motion sequence, the model is trained by minimizing the Mean Squared Error (MSE) between the decoder outputs \(\hat{P}_T=(\hat{p}_1,...,\hat{p}_T)\) and the corresponding 3D ground-truth \(P_T=(p_1,...,p_T)\):
\begin{equation}
L_{MSE} = \sum_{t=1}^{T} \sum_{v=1}^{V} |\hat{p}_{t,v} - p_{t,v}|^2,
\end{equation} 

\noindent
where \(V\) represents the number of vertices in the 3D facial mesh~\cite{fan2022faceformer}. 
To reduce jitter in the output frames caused by relying solely on the MSE loss, we add a velocity loss that promotes smoother and more realistic lip motion over time~\cite{Cudeiro2019Capture}. 
This loss is formulated as follows:
\begin{equation}
L_{VEL} = \sum_{t=1}^{T} \sum_{v=1}^{V} |(\hat{p}_{t,v} - \hat{p}_{t-1,v} ) - (p_{t,v} - p_{t-1,v} )|^2.
\end{equation} 

\begin{figure}[!t]
\includegraphics[width=14cm]{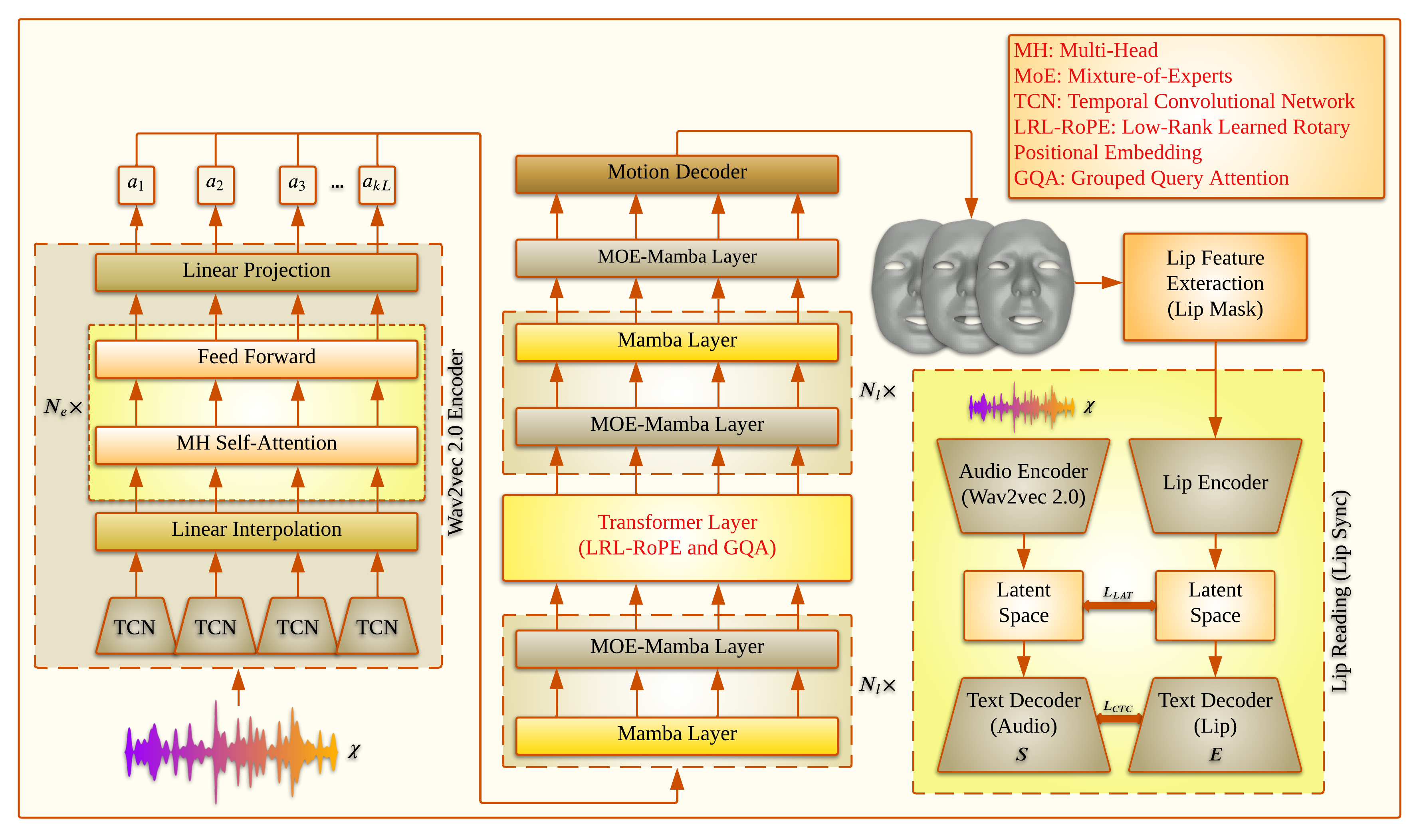}
\Description{}
\centering
\caption[]{\textbf{Overview of JambaTalk:} The Wav2Vec~2.0 model is used to extract features from the input speech, with the encoder initialized using pre-trained weights from the original model~\cite{Baevski2020wav2vec}. These encoded features are passed to the JambaTalk decoder, which generates a sequence of animated 3D face meshes. The Transformer layer incorporates Low-Rank Learned Rotary Positional Embedding (LRL-RoPE) and Grouped-Query Attention (GQA), providing a computation-efficient alternative to traditional Transformers. The lip feature extraction block then converts motion decoder outputs into lip deformation features by selecting lip vertices with a lip mask, which are processed by a Transformer-based lip encoder in the lip reader module to synchronize lip shapes.}
\label{fig:fig1}
\end{figure}

\noindent
As depicted in Figure~\ref{fig:fig1}, the lip feature extraction block transforms mesh decoder outputs into lip deformation features by selecting lip vertices using a lip mask. These are processed by a Transformer-based lip encoder to obtain latent lip representations in the lip reader module. In parallel, the pre-trained wav2vec~2.0 model extracts rich latent features from audio and generates text through a pre-trained language decoder, which maps the features to a 32-character vocabulary plus a blank token, selects the highest probability outputs, and applies post-processing to form the transcription. This produces both latent and text features without requiring ground-truth text input. The lip latent features are aligned with audio-derived features using a latent consistency loss, and a text decoder maps the lip features to the same vocabulary space. The resulting outputs are compared with the audio transcriptions using CTC loss to ensure consistency~\cite{Peng2023Selftalk}. The formula for calculating the latent consistency loss is:
\begin{equation}
L_{LAT} = \frac{1}{T} \sum_{t=1}^{T} \frac{1}{N} \sum_{i=1}^{N} \frac{( als_{t,n} - lls_{t,n})^2}{2},
\end{equation} 

\noindent
where \(als_{t,n}\) represents the features extracted from the audio encoder, while \(lls_{t,n}\) represents the features extracted from the lip encoder. \(N\) denotes the number of samples. The CTC loss can be defined as:
\begin{equation}
L_{\mathrm{CTC}} = - \log p \big( \pi \in \mathcal{B}^{-1}(S) \mid \mathbf{e}_{1:T} \big),
\end{equation}

\noindent
where \(\mathbf{e}_{1:T} = (\mathbf{e}_1, \mathbf{e}_2, \dots, \mathbf{e}_T)\) represents the sequence of lip features extracted from the text decoder, \(S = (s_1, s_2, \dots, s_T)\) is the target transcription, \(\pi\) is a possible alignment sequence between \(\mathbf{e}_{1:T}\) and \(S\). The operator \(\mathcal{B}\) removes repeated labels and blanks, and \(\mathcal{B}^{-1}(S)\) represents all valid alignment sequences corresponding to \(S\).

\noindent
To compute the CTC loss, a projection layer maps the lip features to output probabilities:
\begin{equation}
\mathbf{p}_t = \mathrm{softmax} \big( \mathbf{W}_p \mathbf{e}_t + \mathbf{b}_p \big),
\end{equation}

\noindent
where \(\mathbf{W}_p \in \mathbb{R}^{D \times U}\) and \(\mathbf{b}_p \in \mathbb{R}^U\) are learnable parameters, $D$ is the dimension of the lip feature vector \(\mathbf{e}_t\), and \(U = 33\) is the output vocabulary size \cite{Peng2023Selftalk}. The predicted probability vector \(\mathbf{p}_t\) at each time step is then used in the CTC loss to account for all possible alignments between lip features and the target transcription.

\subsection{Mamba Layer}
\begin{figure}[!t]
\includegraphics[width=14cm]{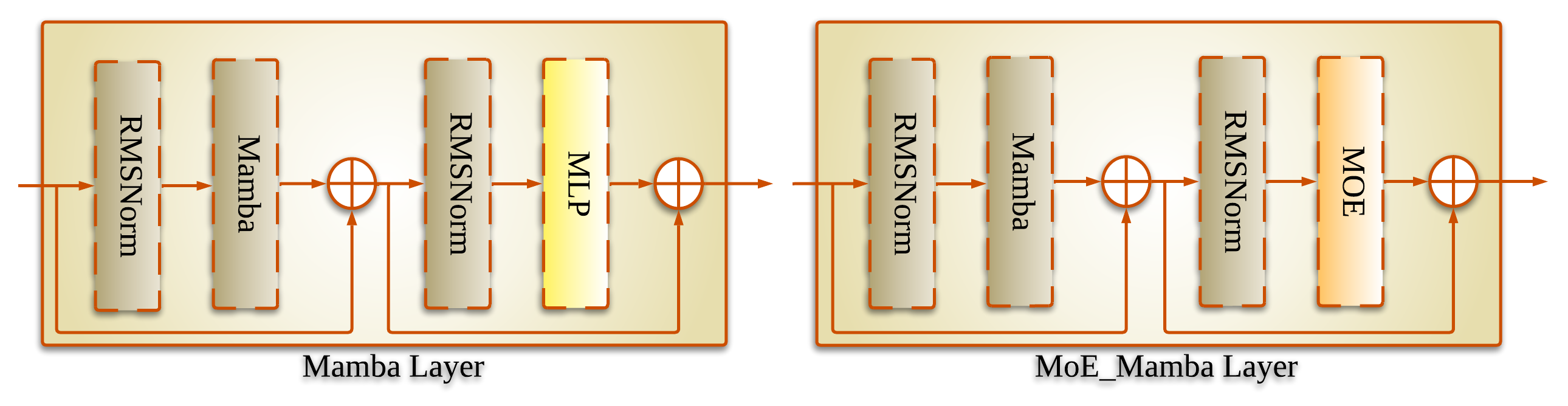}
\Description{}
\centering
\caption[]{Details of the Mamba and MoE\_Mamba layers in the JambaTalk Decoder. Both layers begin with an RMSNorm normalization followed by a Mamba block for sequence modeling and include residual connections to preserve gradient flow. In the standard Mamba Layer (left), the Mamba output is followed by another RMSNorm and a feedforward MLP block. In contrast, the MoE\_Mamba Layer (right) replaces the MLP with a Mixture-of-Experts (MoE) module, enabling dynamic expert routing per token and enhancing model capacity while maintaining computational efficiency~\cite{Lieber2024Jamba}.}
\label{fig:fig2}
\end{figure}

The Scan module (S6) within Mamba~\cite{Gu2023Mamba} is critical in capturing temporal patterns and dependencies in multiple time steps. This module applies a set of trainable operations to each segment of the input sequence, learning key temporal features during training. The parameters governing these operations are optimized to handle long-range dependencies, which makes Mamba particularly effective for generating facial animation, where audiovisual synchronization, fine-grained motion dynamics, and long contextual dependencies must be captured over extended sequences. Figure~\ref{fig:fig2} illustrates the architecture of the Mamba Layer, which consists of two sequential sub-blocks connected by residual pathways. The input first passes through an RMSNorm layer for normalization, followed by a Mamba block that captures long-range dependencies via state-space modeling. The output of the Mamba block is then added to the original input through a residual connection, enhancing gradient flow and stability. This intermediate result undergoes a second RMSNorm and then flows into a standard MLP (multi-layer perceptron) to further transform the features. A second residual connection adds the MLP output back to the intermediate features, forming the final output. The diagram visualizes this modular structure, where normalization, sequence modeling, and feedforward processing are interleaved with skip connections to facilitate deep learning and efficient representation learning.

Mamba improves upon prior state space models~\cite{Gu2022On} like Structured State Space Model (S4)~\cite{Gu2022S4nd} by introducing a selective mechanism that dynamically generates the state parameters \( A \), \( B \), and \( C \) from the input sequence. This dynamic generation is realized through fully connected layers that take the current input and produce the parameters as time-varying functions rather than fixed matrices, enabling adaptive modeling of various motion and speech patterns. Mamba employs the parallel scan algorithm, which leverages the associativity of operations like matrix multiplication to split the input into chunks, process them in parallel, and then merge the results efficiently, achieving parallelism comparable to convolution. The resulting Selective Scan combines the adaptability of dynamic recurrence with the speed of parallel computation. It selectively updates internal memory states based on input, enabling long-range reasoning with input-awareness similar to attention mechanisms, but without their high computational cost. For each batch and dimension, the model processes the input \(x_t\), the hidden state \(h_t\), and output \(y_t\) at each time step \(t\). The formulation of the model is as follows:
\begin{equation}
\begin{matrix}
h_t=\hat{A}_t h_{t-1} + \hat{B}_t x_t, \\
y_t=C_t h_t, 
\end{matrix}
\end{equation}

\noindent
where \(\hat{A}_t\), \(\hat{B}_t\), and \(C_t\) are matrices and vectors that are updated at each time step, enabling the model to adjust to the temporal dynamics of the input sequence. To discretize the continuous-time dynamics, a sampling interval \( \Delta \) is introduced, and the matrices \(\hat{A}_t\) and \(\hat{B}_t\) are computed as follows:
\begin{equation}
\begin{matrix}
\hat{A}_t=exp(\Delta A_t), \\
\hat{B}_t=(\Delta A_t)^{-1} (exp(\Delta A_t) - I).\Delta B_t ,
\end{matrix}
\end{equation}

\noindent
where \(A_t\) and \(B_t\) are generated by the input-dependent projection layers, and \( I \) is the identity matrix. This approach allows the model to handle variable-length sequences and complex temporal behaviors efficiently \cite{Xu2024Mambatalk}.

Modern GPUs excel at computation but are bottlenecked by the slow transfer of data between small, fast SRAM (close to the processor) and large, slow DRAM (farther away). Deep learning models that repeatedly write intermediate results to DRAM and read them back suffer major slowdowns. Mamba tackles this hardware inefficiency with three key optimizations. First, Kernel Fusion combines multiple sequential operations, such as discretization, selective scanning, and matrix multiplication, into a single fused operation that runs entirely within SRAM, avoiding costly DRAM access. Second, during training, Mamba uses Recomputation, choosing to recalculate intermediate values when needed for backpropagation instead of saving them, which saves time by avoiding DRAM reads. Finally, these techniques are combined with Mamba's dynamic input-sensitive architecture in what are called Selective State Space Models (S6), allowing efficient long-sequence modeling with Transformer-level performance, but significantly reduced memory and compute overhead.

JambaTalk places Mamba layers both before and after the Transformer blocks to capture both local dependencies (via attention) and long-range structures (via SSM). This hybrid layering helps model detailed lip movements and head pose shifts over extended sequences, which are essential for generating high-quality 3D talking head animation.

\subsection{MoE\_Mamba Layer}
JambaTalk employs MoE\_Mamba layers to optimize computational efficiency~\cite{Lieber2024Jamba}. Instead of using all experts in the network for each facial motion, JambaTalk selectively routes data to a subset of experts using a dynamic routing mechanism. The MoE\_Mamba Layer integrates a Mamba block with a Mixture-of-Experts (MoE)~\cite{Fedus2022Switch} mechanism to efficiently model long-range dependencies, while dynamically scaling model capacity as depicted in Figure~\ref{fig:fig2}. First, the input sequence is normalized using RMSNorm and passed through a Mamba block, which captures temporal patterns through a recurrent formulation with learnable convolution and state parameters. A residual connection adds stability by combining the Mamba output with the original input. The result is then normalized again and routed through an MoE layer, where each token is processed by a small, learned subset of experts, enabling the model to specialize across different input patterns. The final output combines the MoE result with the original input via another residual connection, allowing the layer to leverage both sequence modeling and expert specialization in a computationally efficient manner. The MoE routing mechanism operates as follows: given an input \(x\), the router computes logits \(h_{x} = W_{r.x}\), which are converted to gating probabilities via a softmax:
\begin{equation}
p_i(x) = \frac{e^{h(x)_i}}{\sum_j^N e^{h(x)_j}}.
\end{equation}

The top-\(k\) gate values are chosen to route the token \(x\). If \(\tau\) is the set of selected top-\(k\) indices, the layer's output is computed as a linearly weighted combination of each expert's computation on the token, weighted by the gate values:
\begin{equation}
y = \sum_{i\in\tau} p_i(x) E_i(x).
\end{equation}

This approach enables efficient utilization of experts, as only the most relevant ones are engaged for each input. It also preserves model capacity by leveraging the diversity of experts to specialize in different tasks or aspects of the input data.

\subsection{Transformer Layer}
\paragraph{\textbf{Low-Rank Learned Rotary Positional Embedding.}}
In our experiments, Low-Rank Learned Rotary Positional Embedding (LRL-RoPE), an advanced positional embedding technique, is utilized to improve generalization ability for longer sequences. The original RoPE encodes absolute positional information using a rotation matrix, thereby directly integrating relative position dependency into the self-attention mechanism. The method provides several benefits, including adaptability to sequences of varying lengths, a gradual reduction in token inter-dependency as relative distances increase, and incorporating relative position encoding into linear self-attention. 

Given an input vector \(x_i \in \mathbb{R}^d\), where \(d\) is even, instead of applying a rotation to the entire \( d \)-dimensional vector at once, the space is divided into \(d/2\) separate two-dimensional subspaces. Each subspace corresponds to a pair of dimensions, such as \((x_1, x_2), (x_3, x_4), \ldots\), and a rotation is applied independently within each pair. As a result, the overall \( d \times d \) rotary matrix \(R_{\Theta,t}^d\) is constructed as a block diagonal matrix, where each block is a \(2 \times 2\) rotation matrix responsible for rotating the corresponding two-dimensional subspace. The rotary matrix applied at position \(t\) is:
\begin{equation}
R_{\theta,t}^d=\begin{pmatrix} 
cos(t\theta_1) & -sin(t\theta_1) & 0 & 0 & \dots & 0 & 0\\
sin(t\theta_1) & cos(t\theta_1) & 0 \\
0 & 0 & cos(t\theta_2) & -sin(t\theta_2) & \dots & 0 & 0\\
0 & 0 & sin(t\theta_2) & cos(t\theta_2) & \dots & 0 & 0\\
\vdots & \vdots & \vdots & \vdots & \ddots & \vdots & \vdots \\
0 & 0 & 0 & 0 & \dots & cos(t\theta_{d/2}) & -sin(t\theta_{d/2})\\
0 & 0 & 0 & 0 & \dots & sin(t\theta_{d/2}) & cos(t\theta_{d/2}) \\
\end{pmatrix} .
\label{eq:rope}
\end{equation}

The rotation frequency parameters are defined as:
\begin{equation}
\theta_i=10000^{-2(i-1)/d}, i \in [1,2,...,d/2] .
\end{equation}

Different rotation speeds (frequencies) are assigned to each \(2D\) subspace, enabling the model to capture both short- and long-range positional information. 
Each input vector \(x_t\) at position \(t\) is first transformed into two different representations called query and key by multiplying with learned projection matrices \(W_q\) and \(W_k\), respectively. These matrices are trained during model learning to map the input features into spaces suitable for attention computation. Next, the Rotary Positional Embedding (RoPE) is applied by multiplying these query and key vectors with a position-dependent rotation matrix \(R_{\Theta,t}^d\) for queries and \(R_{\Theta,\acute{t}}^d\) for keys. This rotation matrix encodes positional information by rotating each \(2D\) subspace of the vector based on the token position and predefined frequencies, effectively injecting relative position information into the vectors:
\begin{equation}
q_t = R_{\Theta,t}^d W_q x_t, \quad k_{\acute{t}} = R_{\Theta,\acute{t}}^d W_k x_{\acute{t}},
\end{equation}

\noindent
where \(R_{\Theta,t}^d\) and \(R_{\Theta,\acute{t}}^d\) are rotary matrices applied at positions \(t\) and \(\acute{t}\), encoding positional information through rotations. The self-attention dot product between query at position \(t\) and and key at position \(\acute{t}\) is:
\begin{equation}
q_t^\top k_{\acute{t}} = \left( R_{\Theta,t}^d W_q x_t \right)^\top \left( R_{\Theta,\acute{t}}^d W_k x_{\acute{t}} \right) . 
\end{equation}

\noindent
Since the rotary matrices \(R_{\Theta,t}^d\) are orthogonal, their transpose equals their inverse:
\begin{equation}
\left(R_{\Theta,t}^d\right)^\top = \left( R_{\Theta,t}^d \right)^{-1}.
\end{equation}

\noindent
Therefore, the dot product simplifies to:
\begin{equation}
q_t^\top k_{t'} = x_t^\top W_q^\top R_{\Theta, t' - t}^d W_k x_{t'},
\end{equation}

\noindent
where \(R_{\Theta,t'-t}^d = (R_{\Theta,t}^d)^\top R_{\Theta,t'}^d\) incorporates relative positional information through the matrix \(R_{\Theta,\acute{t}-t}^d\) depending only on the position difference \(\acute{t}-t\). This multiplicative encoding preserves stability and allows the model to capture positional relationships more naturally compared to additive embeddings~\cite{Su2024Roformer}.

\begin{algorithm}[H]
    \caption{Low-Rank Learned RoPE (LRL-RoPE)}
    \label{alg:lr-lrope}
    \begin{algorithmic}[1]
    \Require $x \in \mathbb{R}^{B \times T \times d}$, pos\_idx $\in \mathbb{R}^{T}$, $W_1 \in \mathbb{R}^{d \times r}$, $W_2 \in \mathbb{R}^{r \times d}$
    \Function{LowRankLearnedRoPE}{$x$, pos\_idx, $W_1$, $W_2$}
        \State $\theta \gets 1 / (10000^{\text{arange}(0,d,2)/d})$ \Comment{predefined frequencies, size $[d/2]$}
        \State $\text{angles} \gets \text{pos\_idx}[:,\text{None}] \cdot \theta[\text{None},:]$ \Comment{$[T, d/2]$}
        \State $\text{sin\_pos} \gets \sin(\text{angles})$
        \State $\text{cos\_pos} \gets \cos(\text{angles})$
        \State $\text{base\_emb} \gets \text{concat}(\text{sin\_pos}, \text{cos\_pos}, \text{dim=-1})$ \Comment{$[T,d]$}
        \State $\text{learned\_emb} \gets \text{base\_emb} \cdot W_1^\top \cdot W_2^\top$ \Comment{low-rank learnable correction with transposes}
        \State $\Delta \text{sin}, \Delta \text{cos} \gets \text{split}(\text{learned\_emb}, 2)$
        \State $\text{sin\_final} \gets \text{sin\_pos} + \Delta \text{sin}$
        \State $\text{cos\_final} \gets \text{cos\_pos} + \Delta \text{cos}$
        \State $x_{\text{even}} \gets x[..., 0::2]$
        \State $x_{\text{odd}} \gets x[..., 1::2]$
        \State $x_{\text{rot\_even}} \gets x_{\text{even}} * \text{cos\_final}[None, :, :] - x_{\text{odd}} * \text{sin\_final}[None, :, :]$
        \State $x_{\text{rot\_odd}} \gets x_{\text{even}} * \text{sin\_final}[None, :, :] + x_{\text{odd}} * \text{cos\_final}[None, :, :]$
        \State \Return interleave($x_{\text{rot\_even}}, x_{\text{rot\_odd}}$)
    \EndFunction
\end{algorithmic}
\end{algorithm}

In the standard Rotary Position Embedding (RoPE) formulation of~(\ref{eq:rope}), the positional encoding for the \(t\)-th position is defined as a fixed sinusoidal vector, where \(\theta_i\) represents the pre-defined angular frequency for the \(i\)-th channel. These angles are calculated using a closed-form formula and are not updated during training, which could restrict the model's ability to adapt to particular tasks or domains. To address this limitation, we introduce a low-rank learnable correction to the fixed RoPE angles. The approach applies a bottleneck projection to \(R_{\theta,t}^d\), reducing it to a low-dimensional representation:
\begin{equation}
h_t = W_1^\top R_{\theta,t}^d, \quad W_1 \in \mathbb{R}^{d \times r}, \quad r \ll d,
\end{equation}

\noindent
Then, mapping back to the original dimension:
\begin{equation}
\Delta_t = W_2^\top h_t, \quad W_2 \in \mathbb{R}^{r \times d}.
\end{equation}

\noindent
In these last equations, \(W_1 \in \mathbb{R}^{d \times r}\) is a learnable matrix whose transpose, \(W_1^\top\), projects the original \(d\)-dimensional vector into an $r$-dimensional bottleneck, \(h_t\), where \(r \ll d\). This low-dimensional representation captures the most important features of the positional embedding, reducing the number of parameters and computational cost while allowing learning. Similarly, \(W_2 \in \mathbb{R}^{r \times d}\) is a learnable matrix whose transpose, \(W_2^\top\), maps \(h_t\) back to the original \(d\)-dimensional space, producing the final transformed positional embedding $\Delta_t \in \mathbb{R}^{d}$. The two matrices \(W_1\) and \(W_2\) together form a low-rank approximation of a learnable transformation, which makes the positional embeddings more expressive and compatible with downstream layers while keeping the original dimensionality.

The resulting vector \(\Delta_t\) is divided into two halves, \(\Delta \sin_t \in \mathbb{R}^{d/2}\) and \(\Delta \cos_t \in \mathbb{R}^{d/2}\), which serve as learnable offsets for the trigonometric components:
\begin{equation}
\begin{aligned}
\sin_{\text{final}, t} &= \sin(t\theta_d) + \Delta \sin_t, \\
\cos_{\text{final}, t} &= \cos(t\theta_d) + \Delta \cos_t.
\end{aligned}
\end{equation}

The rotary transformation is then applied to each 2\(d\) subspace of the token vector \(x_t \in \mathbb{R}^{d}\):
\begin{equation}
\begin{matrix}
x'_{t,2i} = x_{t,2i} \cdot \cos_{\text{final},t,i} - x_{t,2i+1} \cdot \sin_{\text{final},t,i}, \\
x'_{t,2i+1} = x_{t,2i} \cdot \sin_{\text{final},t,i} + x_{t,2i+1} \cdot \cos_{\text{final},t,i}.
\end{matrix}
\end{equation}

This formulation preserves the RoPE's ability to encode relative positions, while allowing task-specific rotation adjustments. The low-rank factorization constrains the additional parameters to \(O(dr)\), providing flexibility at minimal computational cost. The complete mechanism is summarized in Algorithm~\ref{alg:lr-lrope}.

\paragraph{\textbf{Grouped Query Attention}.}
JambaTalk replaces Multi-head Attention (MHA) \cite{Vaswani2017Attention, Shazeer2019Fast} with Grouped-query Attention (GQA) \cite{Ainslie2023Gqa} in the Transformer block, reducing the number of key and value heads to a more manageable number, which in turn decreases both the size of the key-value cache and the overall amount of data processed during attention calculations. This reduction leads to faster training and inference times, while maintaining performance quality close to that of MHA.

\section{Experiments and Results}
\subsection{Experiments Setting}
\paragraph{\textbf{$\bm{BIWI}$ Dataset.}}
The \(BIWI\) dataset~\cite{fanelli2010a} includes 14 subjects, each tasked with reciting 40 distinct sentences, captured at a frame rate of 25 frames per second. Each sentence is about 5 seconds, and the 3D facial meshes are aligned to a uniform topology to ensure consistency across subjects. 
The \(BIWI_6\)~\cite{nocentini2024scantalk} is a reduced dataset that holds a fixed topology with 3,895 vertices and 7,539 faces, while preserving sufficient detail for facial motion analysis. Utilizing this dataset enables more efficient processing without significantly compromising the accuracy of the models, making it well-suited for experiments, where computational resources are a limiting factor. The data set is divided into several subsets: a training set (\(BIWI_6\)-Train) comprising 191 sentences spoken by six subjects (32 sentences per subject) and a validation set (\(BIWI_6\)-Val) consisting of 24 sentences from the same six subjects (4 sentences each).

\paragraph{\textbf{Vocaset Dataset.}}
We employed the publicly accessible 3D dataset, VOCASET~\cite{Cudeiro2019Capture}, for training and testing. This dataset includes 480 speech sequences derived from 3D face scans, showcasing the 12 identities found in the CoMA dataset~\cite{Ranjan2018Generating}. Each 3D face mesh comprises 5,023 vertices. To provide a fair comparison, we employ the same training, validation, and testing splits as VOCA~\cite{Cudeiro2019Capture}, referred to as VOCA-Train, VOCA-Val, and VOCA-Test, respectively. The dataset provides paired audio and 3D facial scans of English utterances. VOCASET includes 255 unique sentences, with some sentences shared across different speakers.

\paragraph{\textbf{Baseline Methods.}}
We compare JambaTalk with four state-of-the-art methods, FaceFormer~\cite{{fan2022faceformer}}, CodeTalker~\cite{xing2023codetalker}, FaceDiffuser~\cite{Stan2023Facediffuser}, and SelfTalk~\cite{Peng2023Selftalk}. We employ the official implementations of these models for training and testing on the Vocaset and \(BIWI_6\) datasets.

\paragraph{\textbf{Implementation Details.}}
We used the Adam optimizer with a learning rate of \(10^{-4}\) for Vocaset, and \(10^{-5}\) for \(BIWI_6\). All model training was conducted on a Linux system with an ASUS NVIDIA GeForce RTX 4090 GPU, and the procedure took approximately 90 minutes (100 epochs) to complete. The encoder parameters are initialized with pre-trained Wav2Vec 2.0 weights. During training, only the TCN parameters remain fixed.

\subsection{Quantitative Comparison}
Following FaceFormer~\cite{fan2022faceformer}, CodeTalker~\cite{xing2023codetalker}, FaceDiffuser~\cite{Stan2023Facediffuser}, and SelfTalk~\cite{Peng2023Selftalk}, we utilize the Mean Vertex Error (MVE) to measure 3D facial mesh reconstruction accuracy over time, the Lip Vertex Error (LVE) to assess lip synchronization, and upper-face motion statistics to evaluate overall facial dynamics, as publicly available metrics to evaluate speech-driven facial animation. Additionally, inspired by ARTalk~\cite{Chu2025ARTalk} and StyleSpeaker~\cite{Yang2025StyleSpeaker}, we employ the Fourier Frequency Error (FFE) to analyze the frequency domain characteristics of motion, and the Motion Offset Deviation (MOE) to assess the smoothness of frame-to-frame transitions. To quantify changes in motion speed, we further use the Acceleration Error (AE).

\paragraph{\textbf{Mean Vertex Error}}
The Mean Vertex Error (MVE) measures the average Euclidean distance between predicted and ground-truth 3D facial vertices across all frames in a sequence. It serves as an indicator of how accurately the model reconstructs the 3D facial mesh over time. A lower MVE value signifies better performance, with values closer to zero indicating that the predicted mesh aligns closely with the true facial geometry. This metric is widely used to assess the overall quality and precision of 3D face reconstruction models~\cite{Stan2023Facediffuser}.

\paragraph{\textbf{Lip Vertex Error}} 
This metric measures the deviation of the lips in a sequence compared to the ground-truth by computing the maximum error \(L2\) of all the lip vertices for each frame and then averaging these values across all frames~\cite{fan2022faceformer, xing2023codetalker, Peng2023Selftalk}.

\paragraph{\textbf{Upper-Face Dynamics Deviation}} 
The Upper-Face Dynamics Deviation (FDD) is only loosely related to speech, as it can vary based on individual speaking styles and the content of the speech. To address this, we measure the variation in facial dynamics of a motion sequence relative to the ground-truth~\cite{xing2023codetalker}.

\paragraph{\textbf{Fourier Frequency Error}} 
In motion synthesis tasks, assessing the accuracy of generated movements is relatively straightforward; however, measuring style consistency remains challenging due to the absence of a universally accepted quantitative metric. Although earlier studies commonly rely on FDD to evaluate style, this approach may fail to capture the unique motion characteristics of different speakers. The Frequency Feature Error (FFE) addresses this limitation by analyzing the frequency domain representation of motion, offering a more precise measure of how well the synthesized motion's rhythmic and frequency patterns align with those of the reference~\cite{Yang2025StyleSpeaker}.

\paragraph{\textbf{Motion Offset Deviation}} 
The Motion Offset Deviation (MOD) evaluates how closely the frame-to-frame transitions in the predicted motion sequence align with those in the ground-truth, effectively capturing the temporal accuracy of the animation. A lower MOD indicates more natural and well-timed motions~\cite{Chu2025ARTalk}.

\paragraph{\textbf{Acceleration Error}} 
The Acceleration Error (AE) quantifies the discrepancy in motion acceleration between predicted and ground-truth sequences, providing a measure of temporal smoothness and realism in facial animation. Let \(\hat{V}_t\) and \(V_t\) denote the predicted and ground-truth vertex positions at time \(t\), respectively. The velocity is computed as the first-order difference: \(\Delta \hat{V}_t = \hat{V}_t - \hat{V}_{t-1}\) and \(\Delta V_t = V_t - V_{t-1}\). The acceleration is then estimated using the second-order difference: \(\hat{a}_t = \Delta \hat{V}_t - \Delta \hat{V}_{t-1}\) and \(a_t = \Delta V_t - \Delta V_{t-1}\). The Acceleration Error is defined as the mean L2 distance between predicted and ground-truth accelerations over all valid time steps:
\begin{equation}
AE = \frac{1}{T - 2} \sum_{t=2}^{T - 1} |\hat{a}_t - a_t|_2 ,
\end{equation}

\noindent
where \(T\) is the total number of frames.

\paragraph{\textbf{Discussion}} 
We compare FaceFormer~\cite{fan2022faceformer}, CodeTalker~\cite{xing2023codetalker}, FaceDiffuser~\cite{Stan2023Facediffuser}, SelfTalk~\cite{Peng2023Selftalk}, and our proposed model, JambaTalk. Evaluation is conducted using six complementary metrics: Mean Vertex Error (MVE), Lip Vertex Error (LVE), Upper-Face Dynamics Deviation (FDD), Fourier Frequency Error (FFE), Motion Offset Deviation (MOD), and Acceleration Error (AE). These metrics capture accuracy, synchronization, dynamic consistency, frequency characteristics, temporal smoothness, and motion realism. Results are reported on the Vocaset and $BIWI_6$ datasets, with scores averaged across all test sequences for fair comparison.

As shown in Tables~\ref{tab:table1} and~\ref{tab:table2}, JambaTalk consistently achieves the best performance in terms of LVE and FDD, reflecting accurate lip synchronization and improved modeling of upper-face dynamics. The reduction in FDD demonstrates that our method better preserves subtle temporal variations in facial expressions, which are strongly tied to both speech content and speaking style. While FDD is known to be a volatile metric, the complementary stability of other measures (MVE, MOD, AE) confirms that our improvements extend to the overall motion quality.  

In addition, JambaTalk attains competitive or superior performance on FFE, indicating stronger alignment with the frequency-domain characteristics of natural motion. On MOD and AE, the results show that our model produces temporally smooth transitions with realistic acceleration profiles, comparable to or better than prior baselines. These findings suggest that incorporating the lip-reading module and our revised training strategy leads to more coherent and naturalistic facial animation across both seen and unseen identities.  

Overall, the results validate that JambaTalk advances the state-of-the-art by achieving consistent improvements across multiple dimensions of speech-driven facial animation, balancing geometric accuracy, temporal smoothness, and expressive dynamics.

\begin{table}[t]
    \caption{Quantitative evaluation results on Vocaset Test.}
    \centering
    \resizebox{\linewidth}{!}{%
    \begin{tabular}{lcccccc}
    \toprule
    Methods & $\downarrow$ MVE & $\downarrow$ LVE & $\downarrow$ FDD & $\downarrow$ FFE & $\downarrow$ MOD & $\downarrow$ AE \\
    & ($\times10^{-1}\text{mm}$) & ($\times10^{-3}\text{mm}$) & ($\times10^{-4}\text{mm}$) & ($\times10^{-5}\text{mm}$) & ($\times10^{-2}\text{mm}$) & ($\times10^{-2}\text{mm}$) \\
    \midrule
    FaceFormer~\cite{fan2022faceformer} & 1.2054 & 1.8134 & 3.2874 & 1.2644 & 2.7175 & 2.2942 \\
    CodeTalker~\cite{xing2023codetalker} & 1.2269 & 1.7626 & 1.9832 & 1.1201 & 2.4736 & 1.9802 \\
    FaceDiffuser~\cite{Stan2023Facediffuser} & 1.2248 & 1.8105 & 2.8076 & 1.1078 & \textbf{2.4301} & \textbf{1.9797}\\
    SelfTalk~\cite{Peng2023Selftalk} & 1.1765 & 1.7809 & 2.8580 & 1.1811 & 2.8153 & 2.3231 \\
    \midrule
    JambaTalk (Ours) & \textbf{1.1751} & \textbf{1.6694} & \textbf{1.7592} & \textbf{1.0657} & 2.7906 & 2.3143\\
    \bottomrule
    \end{tabular}
    }
    \label{tab:table1}
\end{table}

\begin{table}[t]
    \caption{Quantitative evaluation results on \textit{$BIWI_6$} Test-B.}
    \centering
    \resizebox{\linewidth}{!}{%
    \begin{tabular}{lcccccc}
    \toprule
    Methods & $\downarrow$ MVE & $\downarrow$ LVE & $\downarrow$ FDD & $\downarrow$ FFE & $\downarrow$ MOD & $\downarrow$ AE \\
    & ($\times10^{-1}\,\text{mm}$) & ($\times10^{-2}\,\text{mm}$) & ($\times10^{-3}\,\text{mm}$) & ($\times10^{-4}\,\text{mm}$) & ($\times10^{-2}\,\text{mm}$) & ($\times10^{-2}\,\text{mm}$) \\
    \midrule
    FaceFormer~\cite{fan2022faceformer} & 1.8254 & 1.0562 & 1.4988 & 1.7596 & 5.1035 & 5.7658 \\
    CodeTalker~\cite{xing2023codetalker} & 1.8109 & 1.0417 & 1.1456 & 1.7974 & 5.2682 & 5.8884 \\
    FaceDiffuser~\cite{Stan2023Facediffuser} & 1.8771 & 1.0799 & 1.1306 & 1.8564 & 5.1173 & 5.6455 \\
    SelfTalk~\cite{Peng2023Selftalk} & 1.7559 & 0.9961 & 0.7523 & 1.5662 & 4.9813 & 5.6823 \\
    \midrule
    JambaTalk (Ours) & \textbf{1.7341} & \textbf{0.9816} & \textbf{0.6315} & \textbf{1.4421} & \textbf{4.9565} & \textbf{5.5664} \\
    \bottomrule
    \end{tabular}
    }
    \label{tab:table2}
\end{table}

\subsection{Qualitative Evaluation}
\begin{figure}[!t]
    \includegraphics[width=\linewidth]{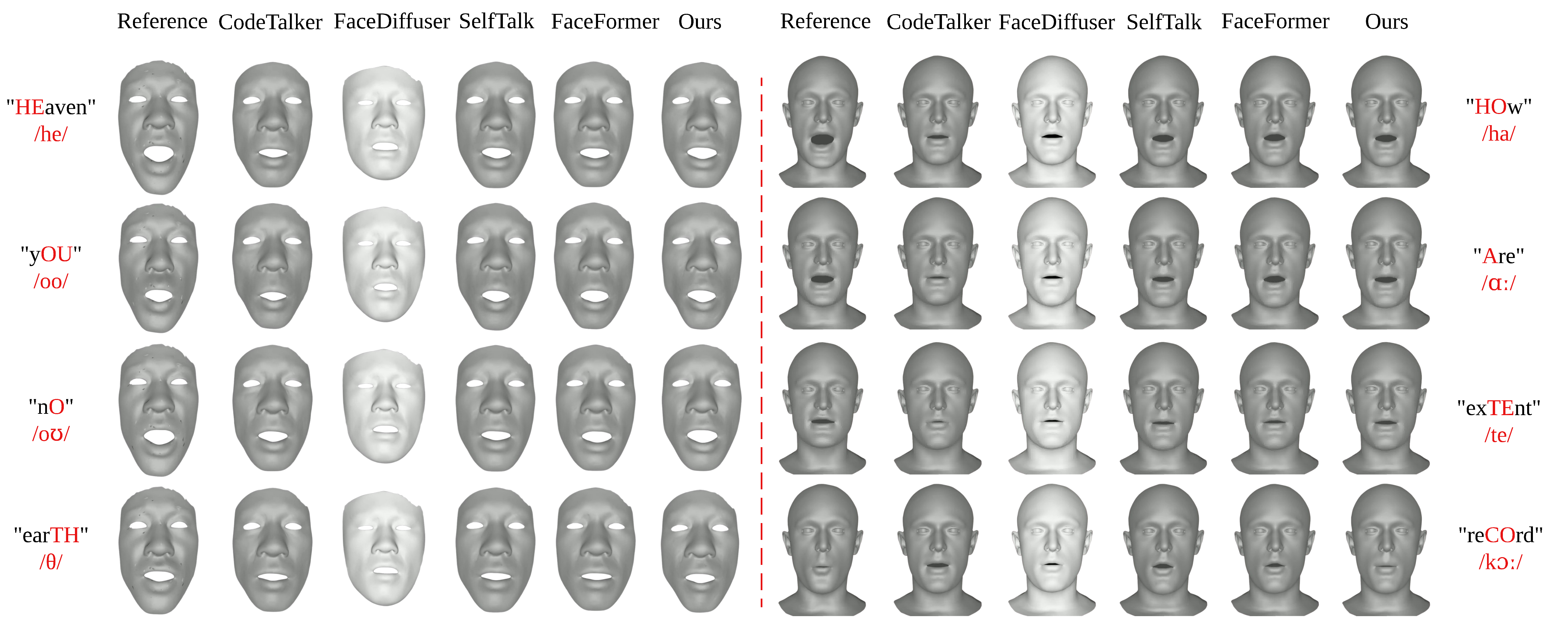}
    \Description{}
    \centering
    \caption{A visual comparison of frames from synthesized facial animation sequences produced by various methods, alongside reference frames from the ground-truth sequence. The red utterances are depicted in the visual frames. Our approach generates lip shapes that closely resemble the reference frames. Left: \textit{$BIWI_6$} Test-B. Right: Vocaset Test.} 
    \label{fig:fig3}
\end{figure}

We provide a visual comparison of our method against other competitors. To ensure a fair comparison, we use the same talking style as conditional input for FaceFormer, %\cite{fan2022faceformer}, 
CodeTalker, %\cite{xing2023codetalker}, 
FaceDiffuser, %\cite{Stan2023Facediffuser}, 
SelfTalk, %\cite{Peng2023Selftalk}, 
and our JambaTalk, with the style randomly sampled. To assess lip synchronization performance, we show representative frames of the generated facial animations, each corresponding to specific syllables, as shown in Figure~\ref{fig:fig3}.

Our analysis demonstrates that the lip movements generated by JambaTalk are more precisely synchronized with the speech signal and closely match the reference video (ground-truth). As illustrated in Figure \ref{fig:fig3}, JambaTalk produces accurate lip shapes for different phonemes, including challenging visemes such as \textit{/he/}, \textipa{/kO:/}, and \textit{/\texttheta/}. Compared to competing methods, which often suffer from over-smoothed or ambiguous mouth movements, JambaTalk consistently captures distinct articulation patterns with appropriate mouth openings and closures. For example, when pronouncing words such as ``reCOrd'' \textipa{/kO:/} and ``earTH'' (\textit{/\texttheta/}), our model generates sharper and more natural lip motions that align better with the audio cues. Similarly, in sentences containing complex phoneme transitions (e.g., ``exTEnt'' and ``HEaven''), JambaTalk maintains synchronization, while preserving expressive dynamics. These results highlight the model's ability to handle fine-grained speech--lip correspondence, supporting the quantitative improvements reported in Tables~\ref{tab:table1} and~\ref{tab:table2}.

\begin{figure}[!t]
    \includegraphics[width=\linewidth]{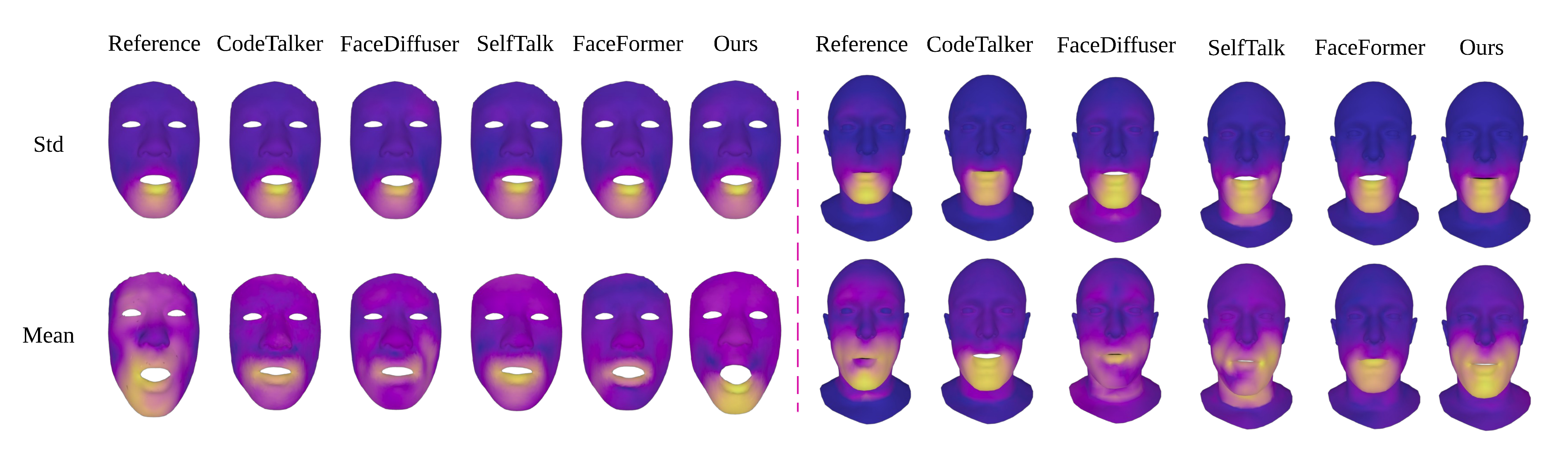}
    \Description{}
    \centering
    \caption{The temporal statistics (mean and standard deviation) of motion variations between adjacent frames in the sequence on Vocaset Test and \textit{$BIWI_6$} Test-B datasets.}
    \label{fig:fig4}
\end{figure}

\begin{figure}[!t]
  \includegraphics[width=\linewidth]{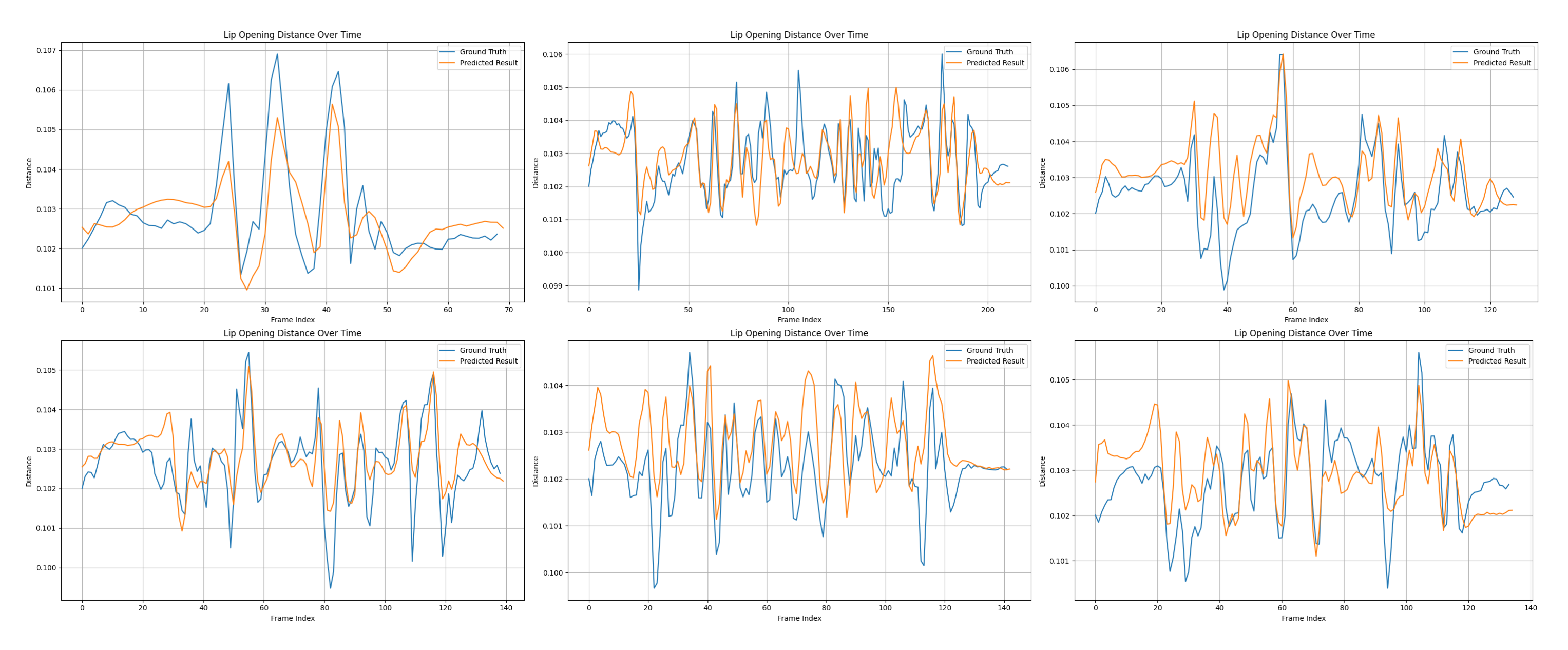}
  \Description{}
  \centering
  \caption{Lip opening distance over time, showing the variation in 3D Euclidean distance between the upper and lower lip landmarks for each video frame. Peaks indicate moments when the mouth is open wider, while valleys correspond to smaller openings or closed lips on \textit{$BIWI_6$} Test-B dataset.}
  \label{fig:fig5}
\end{figure}

Figure~\ref{fig:fig4} presents heatmap visualizations of the mean and standard deviation (Std) errors across all test sequences for different models. The top row shows the per-vertex standard deviation, which reflects the stability and temporal consistency of generated motions, while the bottom row shows the mean error, which highlights the overall accuracy of the reconstructed facial dynamics. Compared to baseline methods, JambaTalk produces heatmaps that are both closer to the reference and exhibit lower intensity in error-prone regions such as the lips and jaw. Specifically, our model shows reduced variance in the mouth and chin areas, indicating more stable motion trajectories without jitter or overshooting. In terms of mean error, JambaTalk yields smaller deviations around the lips and lower face, suggesting more accurate articulation and lip synchronization with the speech signal. By contrast, competing models like CodeTalker and FaceDiffuser suffer from over-smoothing, while SelfTalk and FaceFormer exhibit higher localized errors, particularly around the mouth. These results confirm that JambaTalk not only improves average reconstruction accuracy but also ensures temporally consistent facial motion, leading to more natural and reliable talking-face synthesis.

Based on the plot in Figure~\ref{fig:fig5}, the “Distance” on the Y-axis shows how wide the mouth is open at each moment, measured as the straight-line 3D gap between the upper-lip point and the lower-lip point. When the curve rises, it means the lips are farther apart, indicating the mouth is opening, while dips mean the lips are closer together or touching. The X-axis is the frame index (time), so the plot essentially tracks the mouth's opening and closing pattern throughout the video sequence. Each subplot corresponds to a different test video sequence from the \(BIWI_6\) Test-B dataset.

\begin{figure}
    \includegraphics[width=10cm]{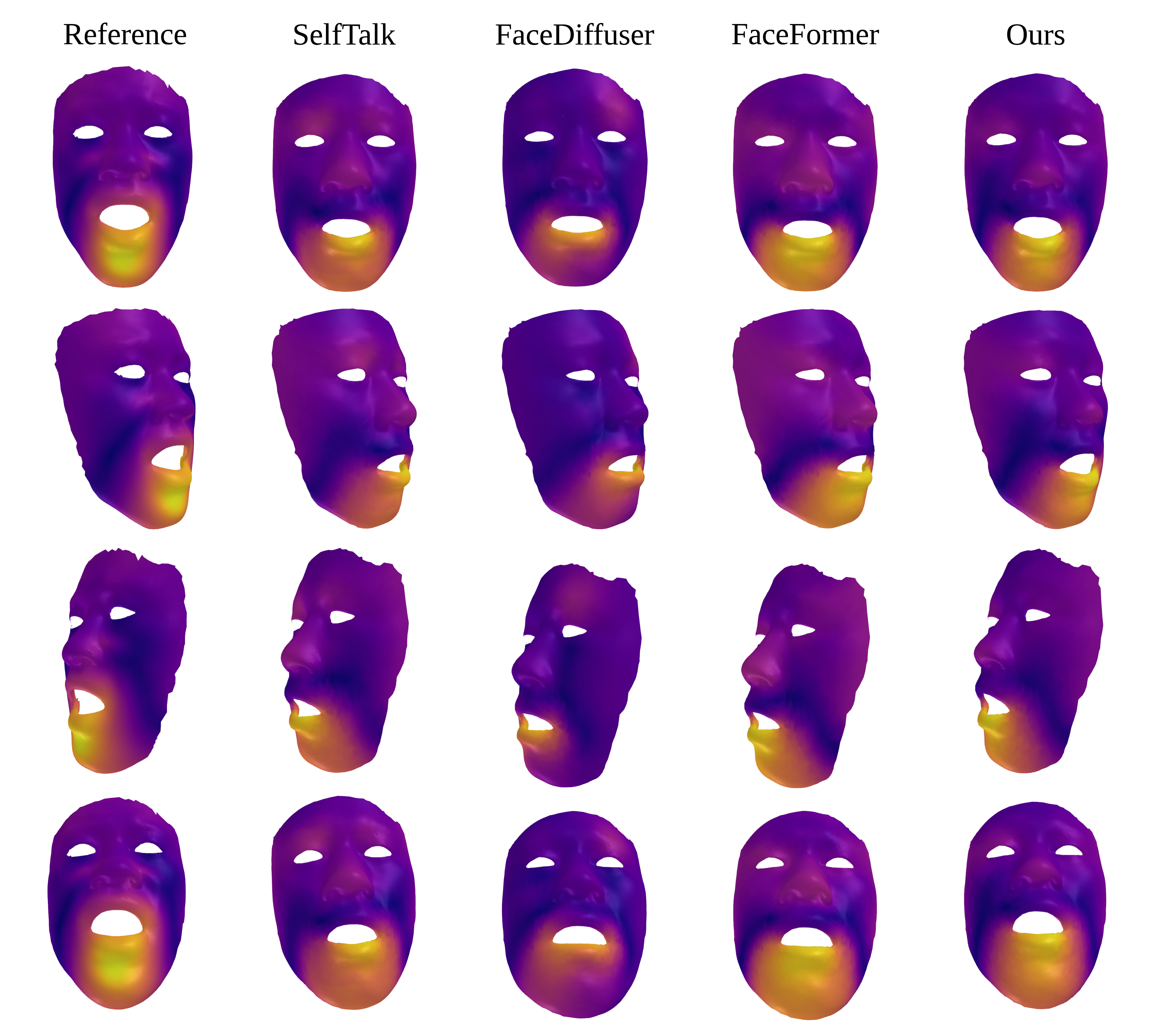}
    \Description{}
    \centering
    \caption{Qualitative comparison of mouth dynamics between the reference and different models (SelfTalk, FaceDiffuser, FaceFormer, and our proposed JambaTalk). The mean heatmaps visualize the lip and jaw motion intensity during speech.}
    \label{fig:fig6}
\end{figure}

Figure~\ref{fig:fig6} shows mean heatmap visualizations of lip and jaw articulation for several representative frames. Compared to baselines, JambaTalk generates lip movements that are both sharper and more consistent with the reference frame. For example, SelfTalk and FaceDiffuser tend to produce smoothed lip shapes, which weakens phoneme articulation, while FaceFormer exhibits exaggerated or unstable mouth openings. By contrast, JambaTalk captures fine-grained details such as the degree of jaw opening and lip closure, resulting in more natural synchronization with the speech signal. These visual results align with our quantitative findings, confirming that JambaTalk achieves more precise and expressive lip–speech correspondence. 

On a single NVIDIA GeForce RTX 4090 (24 GB), JambaTalk can process audio-driven sequences up to 60 seconds in a single pass. Transformer-based baselines such as FaceFormer~\cite{fan2022faceformer} and CodeTalker~\cite{xing2023codetalker} cannot process sequences of this length due to the quadratic memory cost of attention mechanisms. SelfTalk~\cite{Peng2023Selftalk} can handle long sequences similar to JambaTalk but exhibits noticeable quality degradation, including jitter during silent segments, whereas JambaTalk maintains stable and smooth performance (see supplementary video at \href{https://farzanehjafari1987.github.io/JambaTalk.github.io/}{https://JambaTalk.github.io/}).

Diffusion-based methods such as DiffPoseTalk \cite{Sun2024Diffposetalk} can handle sequences longer than those demonstrated in our experiments (up to 2 minutes), but rely on multi-step denoising and sliding-window schemes that involve different computational trade-offs. Our work specifically targets improving the long-sequence capacity of Transformer-style models through the linear-complexity Mamba component.

\section{Ablation Study}
\begin{figure}[!t]
    \includegraphics[width=\linewidth]{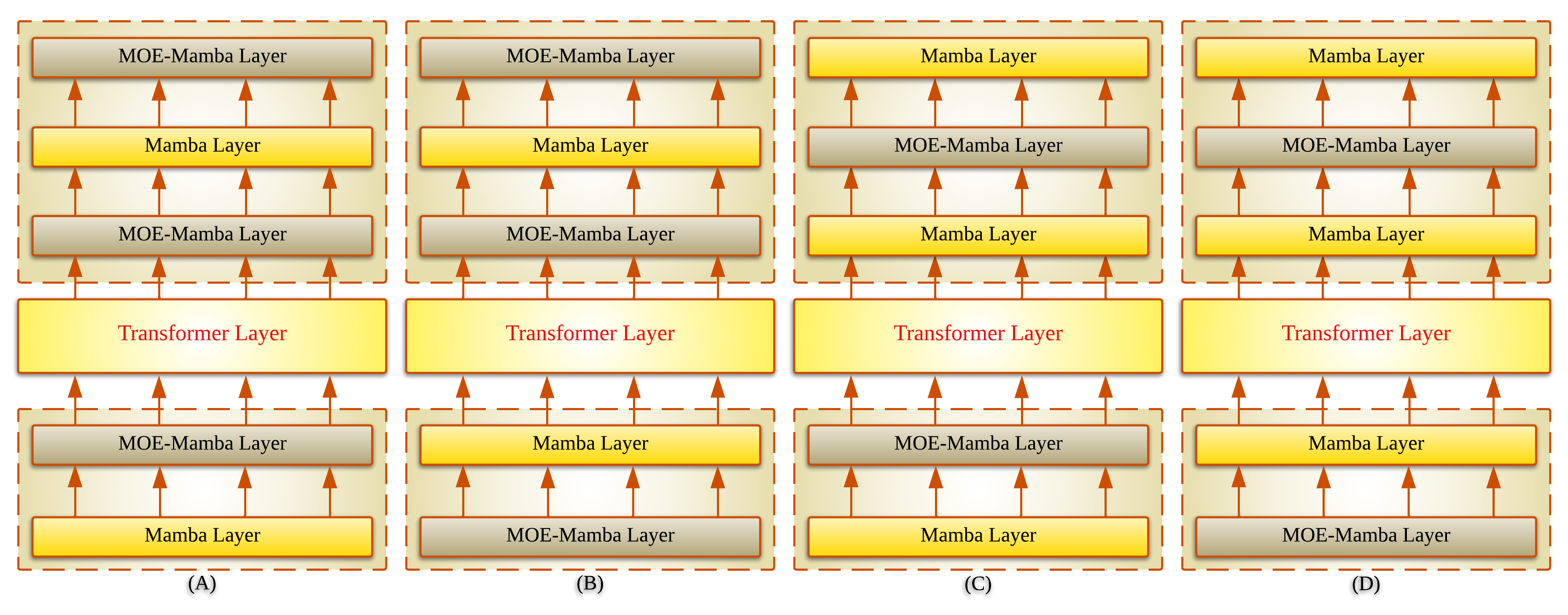}
    \Description{}
    \centering
    \caption{Different sequence of Mamba and MoE\_Mamba layers in the JambaTalk Decoder before and after the Transformer layer. 
    (a) The JambaTalk$_{M-MoE}$ Decoder represents the JambaTalk Decoder with the Mamba layer as the first layer on the left side and the MoE\_Mamba layer as the first layer on the right side of the Transformer; 
    (b) The JambaTalk$_{MoE-MoE}$ Decoder has the MoE\_Mamba layer as the first layer on both sides of the Transformer layer. (c) The JambaTalk$_{M-M}$ Decoder represents the JambaTalk Decoder with the Mamba layers as the first layer on both sides of the Transformer layer. In contrast, in (d), the JambaTalk$_{MoE-M}$ Decoder has the MoE\_Mamba layer as the first layer on the left and the Mamba layer as the first layer on the right side of the Transformer layer.}
    \label{fig:fig7}
\end{figure}

To analyze the effects of the various components in our proposed architecture, we experimented with different configurations by either removing or modifying certain layers. Figure~\ref{fig:fig7} demonstrates different sequences of Mamba and MoE\_Mamba layers in the JambaTalk Decoder before and after the Transformer layer. 
The JambaTalk$_{M-MoE}$ Decoder represents the JambaTalk Decoder with the Mamba layer as the first layer on the left side and the  MoE\_Mamba layer as the first layer on the right side of the Transformer; the  JambaTalk$_{MoE-MoE}$ Decoder has the MoE\_Mamba layer as the first layer on both sides of the Transformer layer. 
The JambaTalk$_{M-M}$ Decoder represents the JambaTalk Decoder with Mamba layers as the first layer on both sides of the Transformer layer. In contrast, the JambaTalk$_{MoE-M}$ Decoder has the MoE\_Mamba layer as the first layer on the left and the Mamba layer as the first layer on the right side of the Transformer. Table~\ref{tab:table3} represents the results of different arrangements of Mamba and MoE\_Mamba layers in the JambaTalk Decoder. Considering all the metrics mentioned, the results of JambaTalk$_{M-MoE}$ are more favorable compared to other sequences.

Table~\ref{tab:table4} presents the performance of the JambaTalk model on the \(BIWI_6\) dataset, comparing configurations with and without Low-Rank Learned Rotary Positional Embedding (LRL-RoPE) and Grouped Query Attention (GQA). The results indicate that JambaTalk achieves better performance when both modules are included, as opposed to when either one or both are excluded. When both components are removed, all metrics degrade, most notably AE (from 5.5664 to 5.7779) and MOD (from 4.9565 to 5.0966), indicating a loss of articulation accuracy and increased motion offset. Using LRL-RoPE alone improves nearly all metrics relative to the no-RoPE/no-GQA baseline, except FDD. Similarly, GQA alone yields a strong MOD (4.9136) and the lowest AE (5.5064) among the single-component variants, reflecting more efficient and accurate attention aggregation. The best performance across nearly all metrics is obtained when LRL-RoPE and GQA are used together (e.g., lowest MVE = 1.7341, LVE = 0.9816, FDD = 0.6315, FFE = 1.4421), showing that they provide complementary benefits for both motion accuracy and visual fidelity.

As shown in Table~\ref{tab:table5}, removing the MoE\_Mamba layers leads to a consistent performance drop across all metrics, indicating their importance for both temporal motion accuracy and facial feature fidelity. Eliminating all Mamba layers yields a slightly lower MVE but significantly worsens other criteria, suggesting that while coarse motion alignment may improve, fine spatial detail suffers as a result. This points to complementary roles: MoE modules capture expressive variation, whereas Mamba layers stabilize temporal consistency. The Lip Reading module and Velocity Loss each provide smaller yet consistent benefits, with their removal increasing errors across most metrics, indicating their complementary roles in improving articulation accuracy and temporal smoothness. 

Table~\ref{tab:table6} compares the model's performance with a lip-reading module against an alternative configuration using motion prior and style embeddings. Incorporating lip reading substantially improves the results, with notable reductions in LVE (from 1.0329 to 0.9816) and FFE (from 1.5640 to 1.4421), indicating more accurate lip motion, enhanced facial feature fidelity, and smoother overall animation. 

We did not utilize both the motion prior and style embedding, as well as the lip-reading module, simultaneously because the lip-reading module alone provides strong phoneme-to-motion alignment, which overlaps with some of the temporal dynamics captured by the motion prior. Combining both would increase model complexity and the number of parameters, potentially leading to overfitting or optimization difficulties, especially given the limited size of available datasets. Moreover, it would add inference overhead without clear empirical benefit, as shown in Table~\ref{tab:table6}, where the lip-reading module alone outperforms the configuration with motion prior and style embedding. Therefore, we opted for the simpler and more effective design that achieves better motion and articulation accuracy, while maintaining computational efficiency.

\begin{table}[t]
    \caption{Quantitative evaluation of different sequences of Mamba and MoE\_Mamba layers in the JambaTalk decoder before and after the Transformer layer on \textit{$BIWI_6$} Test_B.}
    \centering
    \resizebox{\linewidth}{!}{%
    \begin{tabular}{lcccccc}
    \toprule
    Methods & $\downarrow$ MVE & $\downarrow$ LVE & $\downarrow$ FDD & $\downarrow$ FFE & $\downarrow$ MOD & $\downarrow$ AE \\
    & ($\times10^{-1}\,\text{mm}$) & ($\times10^{-2}\,\text{mm}$) & ($\times10^{-3}\,\text{mm}$) & ($\times10^{-4}\,\text{mm}$) & ($\times10^{-2}\,\text{mm}$) & ($\times10^{-2}\,\text{mm}$) \\
    \midrule
    JambaTalk$_{M-MoE}$ & \textbf{1.7341} & \textbf{0.9816} & \underline{0.6315} & \textbf{1.4421} & \textbf{4.9565} & \textbf{5.5664}\\
    JambaTalk$_{MoE-MoE}$ & \underline{1.7985} & \underline{1.0298} & 0.8597 & \underline{1.6769} & \underline{5.0584} & \underline{5.6583}\\
    \midrule
    JambaTalk$_{M-M}$ & 1.8254 & 1.0524 & \textbf{0.5459} & 1.8061 & 5.1640 & 5.7142\\
    JambaTalk$_{MoE-M}$ & 1.8563 & 1.0725 & 0.7610 & 1.8153 & 5.1407 & 5.7058\\
    \bottomrule
    \end{tabular}
    }
    \label{tab:table3}
\end{table}

\begin{table}[t]
  \caption{Ablation study of the JambaTalk model with and without Low-Rank Learned Rotary Positional Embedding (LRL-RoPE) and Grouped Query Attention (GQA) on \textit{$BIWI_6$} Test_B.}
  \centering
  \resizebox{\linewidth}{!}{%
  \begin{tabular}{lcccccc}
    \toprule
    Methods & $\downarrow$ MVE & $\downarrow$ LVE & $\downarrow$ FDD & $\downarrow$ FFE & $\downarrow$ MOD & $\downarrow$ AE \\
     & ($\times10^{-1}\,\text{mm}$) & ($\times10^{-2}\,\text{mm}$) & ($\times10^{-3}\,\text{mm}$) & ($\times10^{-4}\,\text{mm}$) & ($\times10^{-2}\,\text{mm}$) & ($\times10^{-2}\,\text{mm}$) \\
    \midrule
    JambaTalk$_{M-MoE}$ $w/$  LRL-RoPE & 1.7384 & 0.9939 & 0.7802 & 1.5282 & 4.9358 & 5.5471\\
    JambaTalk$_{M-MoE}$ $w/$  GQA \cite{Ainslie2023Gqa} & 1.7466 & 0.9936 & 0.6454 & 1.5982 & \textbf{4.9136} & \textbf{5.5064}\\
    JambaTalk$_{M-MoE}$ $w/o$  LRL-RoPE \& GQA \cite{Ainslie2023Gqa} & 1.7567 & 1.0065 & 0.6868 & 1.5458 & 5.0966 & 5.7779\\
    JambaTalk$_{M-MoE}$ $w/$  LRL-RoPE \& GQA \cite{Ainslie2023Gqa} & \textbf{1.7341} & \textbf{0.9816} & \textbf{0.6315} & \textbf{1.4421} & 4.9565 & 5.5664\\
    JambaTalk$_{M-MoE}$ $w/$  RoPE \cite{Su2024Roformer} \& GQA \cite{Ainslie2023Gqa} & 1.7620 & 1.0079 & 6.9520 & 1.6081 & 4.9474 & 5.5395\\
    \bottomrule
  \end{tabular}
  }
  \label{tab:table4}
\end{table}

\begin{table}[t]
  \caption{Objective metrics on the $BIWI_6$ Test-B dataset for ablation experiments examining the removal of different layers.}
  \centering
  \resizebox{\linewidth}{!}{%
  \begin{tabular}{lcccccc}
    \toprule
    Methods & $\downarrow$ MVE & $\downarrow$ LVE & $\downarrow$ FDD & $\downarrow$ FFE & $\downarrow$ MOD & $\downarrow$ AE \\
     & ($\times10^{-1}\,\text{mm}$) & ($\times10^{-2}\,\text{mm}$) & ($\times10^{-3}\,\text{mm}$) & ($\times10^{-4}\,\text{mm}$) & ($\times10^{-2}\,\text{mm}$) & ($\times10^{-2}\,\text{mm}$) \\
    \midrule
    JambaTalk $w/o$ Lip Reading & 1.7704 & 1.0029 &  \underline{0.6858} & \underline{1.5385} & \underline{4.9620} & \textbf{5.5506}\\
    JambaTalk $w/o$ Velocity Loss Function & 1.7894 & 1.0236 & 0.7023 & 1.5669 & 5.1147 & 5.7670\\
    JambaTalk $w/o$ MoE\_Mamba Layers & 1.8115 & 1.0351 & 0.7249 & 1.6988 & 5.1009 & 5.6791\\
    JambaTalk $w/o$ Mamba Layers & \textbf{1.6967} & \underline{0.9882} & 0.7582 & 1.5581 & 4.9724 & 5.7184\\
    JambaTalk (Ours) & \underline{1.7341} & \textbf{0.9816} & \textbf{0.6315} & \textbf{1.4421} & \textbf{4.9565} & \underline{5.5664}\\
    \bottomrule
  \end{tabular}
  }
  \label{tab:table5}
\end{table}

\begin{table}[t]
  \caption{Impact of the lip reading module on motion and articulation accuracy for the $BIWI_6$ dataset.}
  \centering
  \resizebox{\linewidth}{!}{%
  \begin{tabular}{lcccccc}
    \toprule
    Methods & $\downarrow$ MVE & $\downarrow$ LVE & $\downarrow$ FDD & $\downarrow$ FFE & $\downarrow$ MOD & $\downarrow$ AE \\
     & ($\times10^{-1}\,\text{mm}$) & ($\times10^{-2}\,\text{mm}$) & ($\times10^{-3}\,\text{mm}$) & ($\times10^{-4}\,\text{mm}$) & ($\times10^{-2}\,\text{mm}$) & ($\times10^{-2}\,\text{mm}$) \\
    \midrule
    JambaTalk $w/$ Motion Prior \& Style Embedding & 1.8013 & 1.0329 & 0.8584 & 1.5640 & 5.5294 & 6.3751\\
    JambaTalk $w/$ Lip Reading  & \textbf{1.7341} & \textbf{0.9816} & \textbf{0.6315} & \textbf{1.4421} & \textbf{4.9565} & \textbf{5.5664}\\
    \bottomrule
  \end{tabular}
  }
  \label{tab:table6}
\end{table}

\begin{table}[t]
    \caption{Runtime performance metrics for JambaTalk.}
    \centering
    \resizebox{\linewidth}{!}{%
    \begin{tabular}{lcp{8cm}}
    \toprule
    Metric & Range (Typical) & Interpretation \\
    \midrule
    Inference Time (Latency) & 30.6 ms -- 42.7 ms & Time required to process a 1-second audio window; low latency suitable for real-time applications. \\
    Output FPS (Throughput) & $\sim$600 -- 800 FPS & Rate at which the model generates mesh frames. \\
    Speedup (vs. real-time audio) & $\sim$24$\times$ -- 32$\times$ & Frames generated roughly 30× faster than real-time audio playback. \\
    \bottomrule
   \end{tabular}
    }
  \label{tab:table7}
\end{table}

\section{Runtime Efficiency Analysis}
We have implemented a real-time, speech-driven 3D talking face generation pipeline that efficiently handles both microphone and audio file inputs. Audio signals are streamed into a queue in small chunks, enabling continuous processing without blocking the user interface. Each chunk is converted into features using a Wav2Vec2-based processor and cast to half precision (float16) to optimize GPU memory usage. The model processes these features in overlapping windows, allowing low-latency inference while maintaining temporal continuity. This approach supports high-throughput frame generation, with each audio window producing the corresponding 3D mesh vertices for the talking head.

To render the generated sequences, a dedicated off-screen 3D renderer updates the mesh in real time, providing smooth and interactive visualization. Mesh predictions are placed into a secondary queue consumed by the rendering thread, ensuring synchronization between inference and display. This design enables efficient pipelining, allowing JambaTalk to achieve real-time performance on standard GPUs (e.g., NVIDIA RTX 4090) with low latency and significant speedup over real-time audio playback. The system can generate infinite-length sequences using a sliding-window approach. Runtime performance metrics are summarized in Table~\ref{tab:table7}. (see supplementary video at 
\href{https://farzanehjafari1987.github.io/JambaTalk.github.io/}{https://JambaTalk.github.io/}).

\section{User Study}
\begin{figure}
    \includegraphics[width=11cm]{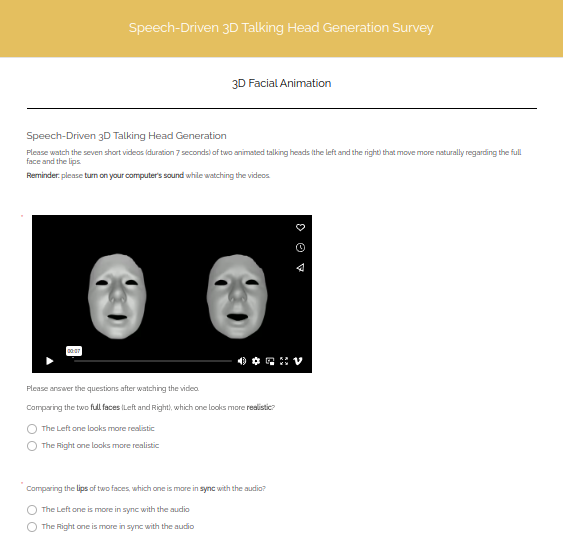}
    \Description{}
    \centering
    \caption{The screenshots of the speech-driven 3D talking head generation user study survey for participants.}
    \label{fig:fig8}
\end{figure}

We conducted a user study using Amazon Mechanical Turk services to evaluate the quality of animated faces in terms of perceptual lip synchronization and realism, comparing our method with FaceFormer~\cite{fan2022faceformer}, CodeTalker~\cite{xing2023codetalker}, FaceDiffuser~\cite{Stan2023Facediffuser}, and SelfTalk~\cite{Peng2023Selftalk}. A user study is a reliable method for evaluating 3D talking faces, so we employed A/B tests for each comparison, pitting our method against competitors in terms of realistic facial animation and lip synchronization (Figure~\ref{fig:fig8}). In this study, 60 participants assessed the animations in side-by-side presentations, choosing the one they considered more realistic based on their preferences. We calculated the ratio of user choices to measure satisfaction. As illustrated in Table~\ref{tab:table8}, our method exhibited superior performance in both perceptual lip synchronization and facial realism. In particular, \(86.4\%\) of the users preferred our realism and \(67.8\%\) favored our lip synchronization on the Vocaset dataset compared to CodeTalker.

\begin{table*}[!h]
    \caption{User study results comparing JambaTalk with baseline models on two criteria—\textit{realism} and \textit{lip synchronization}—across the Vocaset and $BIWI_6$ datasets. Left: comparisons with FaceFormer and CodeTalker. Right: comparisons with FaceDiffuser and SelfTalk.}
     \centering
    \resizebox{\textwidth}{!}{%
    \begin{tabular}{lcccc|lcccc}
    \toprule
    \textbf{Comparison} & \multicolumn{2}{c}{Vocaset Test (\%)} & \multicolumn{2}{c}{$BIWI_6$ Test-B (\%)} &
    \textbf{Comparison} & \multicolumn{2}{c}{Vocaset Test (\%)} & \multicolumn{2}{c}{$BIWI_6$ Test-B (\%)} \\
    \cmidrule(lr){2-3} \cmidrule(lr){4-5} \cmidrule(lr){7-8} \cmidrule(lr){9-10}
    & Competitor & Ours & Competitor & Ours &
    & Competitor & Ours & Competitor & Ours \\
    \midrule
    \textbf{Ours vs. FaceFormer} & & & & &
    \textbf{Ours vs. FaceDiffuser} & & & & \\
    \quad Realism & 30.5 & 69.5 & 39.7 & 60.3 & \quad Realism & 34.5 & 65.5 & 33.9 & 66.1 \\
    \quad Lip Sync & 43.9 & 56.1 & 35.1 & 64.9 & \quad Lip Sync & 32.2 & 67.8 & 40.7 & 59.3 \\
    \midrule
    \textbf{Ours vs. CodeTalker} & & & & &
    \textbf{Ours vs. SelfTalk} & & & & \\
    \quad Realism & 13.6 & 86.4 & 44.1 & 55.9 & \quad Realism & 25.9 & 74.1 & 22.4 & 77.6 \\
    \quad Lip Sync & 32.2 & 67.8 & 48.3 & 51.7& \quad Lip Sync & 39.0 & 61.0 & 35.5 & 64.4 \\
    \bottomrule
    \end{tabular}
    }
    \label{tab:table8}
\end{table*}

\section{Conclusions}
In this work, JambaTalk demonstrated significant progress in predicting facial motion by combining the most acceptable elements of Transformer and Mamba architectures using a mixture-of-experts strategy for speech-driven 3D facial animation. This hybrid design delivered high throughput, efficient memory utilization, and the capacity to manage extensive contexts. This makes our approach a flexible and potent model for a wide range of talking head generation tasks. The quantitative analysis illustrated that our approach outperforms other state-of-the-art methods. However, the overall perceptual quality remains behind the ground-truth, mainly due to the limited availability of paired audio-visual data. In addition, the generic motion prior obtained conforms to the motion patterns defined by the training set, which may differ from actual facial movements in the real world. Although most systems today depend on audio or text prompts, exploring control through modalities such as gestures, gaze, and emotions presents an exciting area for future research.

\begin{acks}
This research was partially funded by the Natural Sciences and Engineering Research Council (NSERC) and the Alberta Innovates Discovery Supplement fund. We thank the authors of VOCA, \(BIWI\), FaceFormer, CodeTalker, FaceDiffuser, SelfTalk, and ScanTalk for providing access to their implementations and datasets.
\end{acks}

%%% -*-BibTeX-*-
%%% Do NOT edit. File created by BibTeX with style
%%% ACM-Reference-Format-Journals [18-Jan-2012].


\begin{thebibliography}{75}

%%% ====================================================================
%%% NOTE TO THE USER: you can override these defaults by providing
%%% customized versions of any of these macros before the \bibliography
%%% command.  Each of them MUST provide its own final punctuation,
%%% except for \shownote{}, \showDOI{}, and \showURL{}.  The latter two
%%% do not use final punctuation, in order to avoid confusing it with
%%% the Web address.
%%%
%%% To suppress output of a particular field, define its macro to expand
%%% to an empty string, or better, \unskip, like this:
%%%
%%% \newcommand{\showDOI}[1]{\unskip}   % LaTeX syntax
%%%
%%% \def \showDOI #1{\unskip}           % plain TeX syntax
%%%
%%% ====================================================================

\ifx \showCODEN    \undefined \def \showCODEN     #1{\unskip}     \fi
\ifx \showDOI      \undefined \def \showDOI       #1{#1}\fi
\ifx \showISBNx    \undefined \def \showISBNx     #1{\unskip}     \fi
\ifx \showISBNxiii \undefined \def \showISBNxiii  #1{\unskip}     \fi
\ifx \showISSN     \undefined \def \showISSN      #1{\unskip}     \fi
\ifx \showLCCN     \undefined \def \showLCCN      #1{\unskip}     \fi
\ifx \shownote     \undefined \def \shownote      #1{#1}          \fi
\ifx \showarticletitle \undefined \def \showarticletitle #1{#1}   \fi
\ifx \showURL      \undefined \def \showURL       {\relax}        \fi
% The following commands are used for tagged output and should be
% invisible to TeX
\providecommand\bibfield[2]{#2}
\providecommand\bibinfo[2]{#2}
\providecommand\natexlab[1]{#1}
\providecommand\showeprint[2][]{arXiv:#2}

\bibitem[Ainslie et~al\mbox{.}(2023)]%
        {Ainslie2023Gqa}
\bibfield{author}{\bibinfo{person}{Joshua Ainslie}, \bibinfo{person}{James Lee-Thorp}, \bibinfo{person}{Michiel de Jong}, \bibinfo{person}{Yury Zemlyanskiy}, \bibinfo{person}{Lebrón Federico}, {and} \bibinfo{person}{Sumit Sanghai}.} \bibinfo{year}{2023}\natexlab{}.
\newblock \showarticletitle{Gqa: Training generalized multi-query transformer models from multi-head checkpoints.}
\newblock \bibinfo{journal}{\emph{arXiv preprint arXiv:2305.13245}} (\bibinfo{year}{2023}).
\newblock


\bibitem[Anderson et~al\mbox{.}(2013)]%
        {anderson2013an}
\bibfield{author}{\bibinfo{person}{Robert Anderson}, \bibinfo{person}{Bjorn Stenger}, \bibinfo{person}{Vincent Wan}, {and} \bibinfo{person}{Roberto Cipolla1}.} \bibinfo{year}{2013}\natexlab{}.
\newblock \showarticletitle{An expressive text-driven 3d talking head.}
\newblock \bibinfo{journal}{\emph{ACM SIGGRAPH 2013 Posters, {1-1}}} (\bibinfo{year}{2013}).
\newblock


\bibitem[Ashish(2017)]%
        {Vaswani2017Attention}
\bibfield{author}{\bibinfo{person}{Vaswani Ashish}.} \bibinfo{year}{2017}\natexlab{}.
\newblock \showarticletitle{Attention is all you need.}
\newblock \bibinfo{journal}{\emph{Advances in neural information processing systems}} \bibinfo{number}{30} (\bibinfo{year}{2017}), \bibinfo{pages}{I}.
\newblock


\bibitem[Azari and Lim(2024)]%
        {Azari2024EmoStyle}
\bibfield{author}{\bibinfo{person}{Bita Azari} {and} \bibinfo{person}{Angelica Lim}.} \bibinfo{year}{2024}\natexlab{}.
\newblock \showarticletitle{EmoStyle: One-Shot Facial Expression Editing Using Continuous Emotion Parameters.}
\newblock \bibinfo{journal}{\emph{Proceedings of the IEEE/CVF Winter Conference on Applications of Computer Vision}} (\bibinfo{year}{2024}).
\newblock


\bibitem[Baevski et~al\mbox{.}(2020)]%
        {Baevski2020wav2vec}
\bibfield{author}{\bibinfo{person}{Alexei Baevski}, \bibinfo{person}{Yuhao Zhou}, \bibinfo{person}{Abdelrahman Mohamed}, {and} \bibinfo{person}{Michael Auli}.} \bibinfo{year}{2020}\natexlab{}.
\newblock \showarticletitle{wav2vec 2.0: A framework for self-supervised learning of speech representations.}
\newblock \bibinfo{journal}{\emph{Advances in neural information processing systems}} \bibinfo{number}{33} (\bibinfo{year}{2020}), \bibinfo{pages}{12449--12460}.
\newblock


\bibitem[Chen et~al\mbox{.}(2023)]%
        {chen2023hyperlips}
\bibfield{author}{\bibinfo{person}{Yaosen Chen}, \bibinfo{person}{Yu Yao}, \bibinfo{person}{Zhiqiang Li}, \bibinfo{person}{Wei Wang}, \bibinfo{person}{Yanru Zhang}, \bibinfo{person}{Han Yang}, {and} \bibinfo{person}{Xuming Wen}.} \bibinfo{year}{2023}\natexlab{}.
\newblock \showarticletitle{HyperLips: Hyper Control Lips with High-Resolution Decoder for Talking Face Generation.}
\newblock \bibinfo{journal}{\emph{arXiv preprint arXiv:2310.05720}} (\bibinfo{year}{2023}).
\newblock


\bibitem[Chu et~al\mbox{.}(2025)]%
        {Chu2025ARTalk}
\bibfield{author}{\bibinfo{person}{Xuangeng Chu}, \bibinfo{person}{Nabarun Goswami}, \bibinfo{person}{Ziteng Cui}, \bibinfo{person}{Hanqin Wang}, {and} \bibinfo{person}{Tatsuya Harada}.} \bibinfo{year}{2025}\natexlab{}.
\newblock \showarticletitle{ARTalk: Speech-Driven 3D Head Animation via Autoregressive Model.}
\newblock \bibinfo{journal}{\emph{arXiv preprint arXiv:2502.20323}} (\bibinfo{year}{2025}).
\newblock


\bibitem[Cudeiro et~al\mbox{.}(2019)]%
        {Cudeiro2019Capture}
\bibfield{author}{\bibinfo{person}{Daniel Cudeiro}, \bibinfo{person}{Timo Bolkart}, \bibinfo{person}{Cassidy Laidlaw}, \bibinfo{person}{Anurag Ranjan}, {and} \bibinfo{person}{Michael J.~Black}.} \bibinfo{year}{2019}\natexlab{}.
\newblock \showarticletitle{Capture, learning, and synthesis of 3D speaking styles.}
\newblock \bibinfo{journal}{\emph{Proceedings of the IEEE/CVF conference on computer vision and pattern recognition}} (\bibinfo{year}{2019}).
\newblock


\bibitem[Daněček et~al\mbox{.}(2023)]%
        {Daněček2023Emotional}
\bibfield{author}{\bibinfo{person}{Radek Daněček}, \bibinfo{person}{Kiran Chhatre}, \bibinfo{person}{Shashank Tripathi}, \bibinfo{person}{Yandong Wen}, \bibinfo{person}{Michael Black}, {and} \bibinfo{person}{Timo Bolkart}.} \bibinfo{year}{2023}\natexlab{}.
\newblock \showarticletitle{Emotional speech-driven animation with content-emotion disentanglement.}
\newblock \bibinfo{journal}{\emph{In SIGGRAPH Asia 2023 Conference Papers}} (\bibinfo{year}{2023}), \bibinfo{pages}{1--13}.
\newblock


\bibitem[Du et~al\mbox{.}(2023)]%
        {Du2023Dae-talker}
\bibfield{author}{\bibinfo{person}{Chenpeng Du}, \bibinfo{person}{Qi Chen}, \bibinfo{person}{Tianyu He}, \bibinfo{person}{Xu Tan}, \bibinfo{person}{Xie Chen}, \bibinfo{person}{Kai Yu}, \bibinfo{person}{Sheng Zhao}, {and} \bibinfo{person}{Jiang Bian}.} \bibinfo{year}{2023}\natexlab{}.
\newblock \showarticletitle{Dae-talker: High fidelity speech-driven talking face generation with diffusion autoencoder.}
\newblock \bibinfo{journal}{\emph{Proceedings of the 31st ACM International Conference on Multimedia}} (\bibinfo{year}{2023}).
\newblock


\bibitem[Fan et~al\mbox{.}(2022)]%
        {fan2022faceformer}
\bibfield{author}{\bibinfo{person}{Yingruo Fan}, \bibinfo{person}{Zhaojiang Lin}, \bibinfo{person}{Jun Saito}, \bibinfo{person}{Wenping Wang}, {and} \bibinfo{person}{Taku Komura}.} \bibinfo{year}{2022}\natexlab{}.
\newblock \showarticletitle{Faceformer: Speech-driven 3d facial animation with transformers.}
\newblock \bibinfo{journal}{\emph{Proceedings of the IEEE/CVF Conference on Computer Vision and Pattern Recognition}} (\bibinfo{year}{2022}).
\newblock


\bibitem[Fanelli et~al\mbox{.}(2010)]%
        {fanelli2010a}
\bibfield{author}{\bibinfo{person}{Gabriele Fanelli}, \bibinfo{person}{Juergen Gall}, \bibinfo{person}{Harald Romsdorfer}, \bibinfo{person}{Thibaut Weise}, {and} \bibinfo{person}{Luc Van~Gool}.} \bibinfo{year}{2010}\natexlab{}.
\newblock \showarticletitle{A 3-D audio-visual corpus of affective communication.}
\newblock \bibinfo{journal}{\emph{IEEE Transactions on Multimedia 12.6}} (\bibinfo{year}{2010}), \bibinfo{pages}{591--598}.
\newblock


\bibitem[Fedus et~al\mbox{.}(2022)]%
        {Fedus2022Switch}
\bibfield{author}{\bibinfo{person}{William Fedus}, \bibinfo{person}{Barret Zoph}, {and} \bibinfo{person}{Noam Shazeer}.} \bibinfo{year}{2022}\natexlab{}.
\newblock \showarticletitle{Switch transformers: Scaling to trillion parameter models with simple and efficient sparsity.}
\newblock \bibinfo{journal}{\emph{Journal of Machine Learning Research 23}} \bibinfo{number}{120} (\bibinfo{year}{2022}), \bibinfo{pages}{1--39}.
\newblock


\bibitem[Gan et~al\mbox{.}(2023)]%
        {gan2023efficient}
\bibfield{author}{\bibinfo{person}{Yuan Gan}, \bibinfo{person}{Zongxin Yang}, \bibinfo{person}{Xihang Yue}, \bibinfo{person}{Lingyun Sun}, {and} \bibinfo{person}{Yi Yang}.} \bibinfo{year}{2023}\natexlab{}.
\newblock \showarticletitle{Efficient emotional adaptation for audio-driven talking-head generation.}
\newblock \bibinfo{journal}{\emph{Proceedings of the IEEE/CVF International Conference on Computer Vision}} (\bibinfo{year}{2023}).
\newblock


\bibitem[Goyal et~al\mbox{.}(2023)]%
        {Goyal2023Emotionally}
\bibfield{author}{\bibinfo{person}{Sahil Goyal}, \bibinfo{person}{Sarthak Bhagat}, \bibinfo{person}{Shagun Uppal}, \bibinfo{person}{Yu~Yi Jangra, Hitkul}, \bibinfo{person}{Yin Yifang}, {and} \bibinfo{person}{Rajiv Ratn~Shah}.} \bibinfo{year}{2023}\natexlab{}.
\newblock \showarticletitle{Emotionally enhanced talking face generation.}
\newblock \bibinfo{journal}{\emph{Proceedings of the 1st International Workshop on Multimedia Content Generation and Evaluation: New Methods and Practice}} (\bibinfo{year}{2023}).
\newblock


\bibitem[Gu and Dao(2023)]%
        {Gu2023Mamba}
\bibfield{author}{\bibinfo{person}{Albert Gu} {and} \bibinfo{person}{Tri Dao}.} \bibinfo{year}{2023}\natexlab{}.
\newblock \showarticletitle{Mamba: Linear-time sequence modeling with selective state spaces.}
\newblock \bibinfo{journal}{\emph{arXiv preprint arXiv:2312.00752}} (\bibinfo{year}{2023}).
\newblock


\bibitem[Gu et~al\mbox{.}(2022)]%
        {Gu2022On}
\bibfield{author}{\bibinfo{person}{Albert Gu}, \bibinfo{person}{Karan Goel}, \bibinfo{person}{Ankit Gupta}, {and} \bibinfo{person}{Christopher Ré}.} \bibinfo{year}{2022}\natexlab{}.
\newblock \showarticletitle{On the parameterization and initialization of diagonal state space models.}
\newblock \bibinfo{journal}{\emph{Advances in Neural Information Processing Systems}} \bibinfo{number}{35} (\bibinfo{year}{2022}), \bibinfo{pages}{35971--35983}.
\newblock


\bibitem[Guan et~al\mbox{.}(2023)]%
        {Guan2023Stylesync}
\bibfield{author}{\bibinfo{person}{Jiazhi Guan}, \bibinfo{person}{Zhou~Hang Zhang, Zhanwang}, \bibinfo{person}{Tianshu Hu}, \bibinfo{person}{Kaisiyuan Wang}, \bibinfo{person}{Dongliang He}, {and} \bibinfo{person}{Haocheng Feng}.} \bibinfo{year}{2023}\natexlab{}.
\newblock \showarticletitle{Stylesync: High-fidelity generalized and personalized lip sync in a style-based generator.}
\newblock \bibinfo{journal}{\emph{Proceedings of the IEEE/CVF Conference on Computer Vision and Pattern Recognition}} (\bibinfo{year}{2023}).
\newblock


\bibitem[Gururani et~al\mbox{.}(2023)]%
        {gururani2023space}
\bibfield{author}{\bibinfo{person}{Siddharth Gururani}, \bibinfo{person}{Arun Mallya}, \bibinfo{person}{Wang Ting-Chun}, \bibinfo{person}{Rafael Valle}, {and} \bibinfo{person}{Ming-Yu Liu}.} \bibinfo{year}{2023}\natexlab{}.
\newblock \showarticletitle{Space: Speech-driven portrait animation with controllable expression.}
\newblock \bibinfo{journal}{\emph{Proceedings of the IEEE/CVF International Conference on Computer Vision}} (\bibinfo{year}{2023}).
\newblock


\bibitem[Hossain and Muhammad(2019)]%
        {Hossain2019Emotion}
\bibfield{author}{\bibinfo{person}{M.~Shamim Hossain} {and} \bibinfo{person}{Ghulam Muhammad}.} \bibinfo{year}{2019}\natexlab{}.
\newblock \showarticletitle{Emotion recognition using deep learning approach from audio-visual emotional big data.}
\newblock \bibinfo{journal}{\emph{Information Fusion 49}} (\bibinfo{year}{2019}), \bibinfo{pages}{69--78}.
\newblock


\bibitem[Huang et~al\mbox{.}(2014)]%
        {Huang2014Speech}
\bibfield{author}{\bibinfo{person}{Zhengwei Huang}, \bibinfo{person}{Ming Dong}, \bibinfo{person}{Qirong Mao}, {and} \bibinfo{person}{Yongzhao Zhan}.} \bibinfo{year}{2014}\natexlab{}.
\newblock \showarticletitle{Speech emotion recognition using CNN.}
\newblock \bibinfo{journal}{\emph{In Proceedings of the 22nd ACM international conference on Multimedia, pp. 801-804.}} (\bibinfo{year}{2014}).
\newblock


\bibitem[Jain et~al\mbox{.}(2020)]%
        {Jain2020Speech}
\bibfield{author}{\bibinfo{person}{Manas Jain}, \bibinfo{person}{Shruthi Narayan}, \bibinfo{person}{Pratibha Balaji}, \bibinfo{person}{Abhijit Bhowmick}, {and} \bibinfo{person}{Rajesh Kumar~Muthu}.} \bibinfo{year}{2020}\natexlab{}.
\newblock \showarticletitle{Speech emotion recognition using support vector machine.}
\newblock \bibinfo{journal}{\emph{arXiv preprint arXiv:2002.07590}} (\bibinfo{year}{2020}).
\newblock


\bibitem[Ji et~al\mbox{.}(2022)]%
        {Ji2022Emma}
\bibfield{author}{\bibinfo{person}{Xinya Ji}, \bibinfo{person}{Hang Zhou}, \bibinfo{person}{Kaisiyuan Wang}, \bibinfo{person}{Qianyi Wu}, \bibinfo{person}{Wayne Wu}, \bibinfo{person}{Feng Xu}, {and} \bibinfo{person}{Xun Cao}.} \bibinfo{year}{2022}\natexlab{}.
\newblock \showarticletitle{Eamm: One-shot emotional talking face via audio-based emotion-aware motion model.}
\newblock \bibinfo{journal}{\emph{In ACM SIGGRAPH 2022 Conference Proceedings}} (\bibinfo{year}{2022}), \bibinfo{pages}{1--10}.
\newblock


\bibitem[Ko et~al\mbox{.}(2024)]%
        {Ko2024Talk3D}
\bibfield{author}{\bibinfo{person}{Jaehoon Ko}, \bibinfo{person}{Kyusun Cho}, \bibinfo{person}{Joungbin Lee}, \bibinfo{person}{Heeji Yoon}, \bibinfo{person}{Sangmin Lee}, \bibinfo{person}{Ahn Sangjun}, {and} \bibinfo{person}{Seungryong Kim}.} \bibinfo{year}{2024}\natexlab{}.
\newblock \showarticletitle{Talk3D: High-Fidelity Talking Portrait Synthesis via Personalized 3D Generative Prior.}
\newblock \bibinfo{journal}{\emph{arXiv preprint arXiv:2403.20153}} (\bibinfo{year}{2024}).
\newblock


\bibitem[Kwon et~al\mbox{.}(2003)]%
        {Kwon2003Emotion}
\bibfield{author}{\bibinfo{person}{Oh-Wook Kwon}, \bibinfo{person}{Kwokleung Chan}, \bibinfo{person}{Jiucang Hao}, {and} \bibinfo{person}{Te-Won Lee}.} \bibinfo{year}{2003}\natexlab{}.
\newblock \showarticletitle{Emotion recognition by speech signals.}
\newblock \bibinfo{journal}{\emph{In Interspeech, pp. 125-128.}} (\bibinfo{year}{2003}).
\newblock


\bibitem[LeCun et~al\mbox{.}(1998)]%
        {LeCun1998Gradient}
\bibfield{author}{\bibinfo{person}{Yann LeCun}, \bibinfo{person}{Léon Bottou}, \bibinfo{person}{Yoshua Bengio}, {and} \bibinfo{person}{Patrick Haffner}.} \bibinfo{year}{1998}\natexlab{}.
\newblock \showarticletitle{Gradient-based learning applied to document recognition.}
\newblock \bibinfo{journal}{\emph{Proceedings of the IEEE 86}} \bibinfo{number}{11} (\bibinfo{year}{1998}), \bibinfo{pages}{2278--2324}.
\newblock


\bibitem[Liang et~al\mbox{.}(2022)]%
        {liang2022expressive}
\bibfield{author}{\bibinfo{person}{Borong Liang}, \bibinfo{person}{Yan Pan}, \bibinfo{person}{Zhizhi Guo}, \bibinfo{person}{Hang Zhou}, \bibinfo{person}{Zhibin Hong}, \bibinfo{person}{Xiaoguang Han}, \bibinfo{person}{Junyu Han}, \bibinfo{person}{Jingtuo Liu}, \bibinfo{person}{Errui Ding}, {and} \bibinfo{person}{Jingdong Wang}.} \bibinfo{year}{2022}\natexlab{}.
\newblock \showarticletitle{Expressive talking head generation with granular audio-visual control.}
\newblock \bibinfo{journal}{\emph{Proceedings of the IEEE/CVF Conference on Computer Vision and Pattern Recognition}} (\bibinfo{year}{2022}).
\newblock


\bibitem[Lieber et~al\mbox{.}(2024)]%
        {Lieber2024Jamba}
\bibfield{author}{\bibinfo{person}{Opher Lieber}, \bibinfo{person}{Bata~Hofit Lenz, Barak}, \bibinfo{person}{Gal Cohen}, \bibinfo{person}{Jhonathan Osin}, \bibinfo{person}{Itay Dalmedigos}, \bibinfo{person}{Erez Safahi}, {and} \bibinfo{person}{et al.}} \bibinfo{year}{2024}\natexlab{}.
\newblock \showarticletitle{Jamba: A hybrid transformer-mamba language model.}
\newblock \bibinfo{journal}{\emph{arXiv preprint arXiv:2403.19887}} (\bibinfo{year}{2024}).
\newblock


\bibitem[Liu et~al\mbox{.}(2023)]%
        {Liu2023Moda}
\bibfield{author}{\bibinfo{person}{Yunfei Liu}, \bibinfo{person}{Lijian Lin}, \bibinfo{person}{Fei Yu}, \bibinfo{person}{Changyin Zhou}, {and} \bibinfo{person}{Yu Li}.} \bibinfo{year}{2023}\natexlab{}.
\newblock \showarticletitle{Moda: Mapping-once audio-driven portrait animation with dual attentions.}
\newblock \bibinfo{journal}{\emph{Proceedings of the IEEE/CVF International Conference on Computer Vision}} (\bibinfo{year}{2023}).
\newblock


\bibitem[Mekruksavanich et~al\mbox{.}(2020)]%
        {Mekruksavanich2020Negative}
\bibfield{author}{\bibinfo{person}{Sakorn Mekruksavanich}, \bibinfo{person}{Anuchit Jitpattanakul}, {and} \bibinfo{person}{Narit Hnoohom}.} \bibinfo{year}{2020}\natexlab{}.
\newblock \showarticletitle{Negative emotion recognition using deep learning for thai language.}
\newblock \bibinfo{journal}{\emph{In 2020 joint international conference on digital arts, media and technology with ECTI northern section conference on electrical, electronics, computer and telecommunications engineering (ECTI DAMT \& NCON), IEEE}} (\bibinfo{year}{2020}), \bibinfo{pages}{71--74}.
\newblock


\bibitem[Nguyen et~al\mbox{.}(2022)]%
        {Gu2022S4nd}
\bibfield{author}{\bibinfo{person}{Eric Nguyen}, \bibinfo{person}{Karan Goel}, \bibinfo{person}{Albert Gu}, \bibinfo{person}{Gordon Downs}, \bibinfo{person}{Preey Shah}, \bibinfo{person}{Tri Dao}, \bibinfo{person}{Stephen Baccus}, {and} \bibinfo{person}{Christopher Ré}.} \bibinfo{year}{2022}\natexlab{}.
\newblock \showarticletitle{S4nd: Modeling images and videos as multidimensional signals with state spaces.}
\newblock \bibinfo{journal}{\emph{Advances in neural information processing systems}} \bibinfo{number}{35} (\bibinfo{year}{2022}), \bibinfo{pages}{2846--2861}.
\newblock


\bibitem[Nocentini and Berretti.(2023)]%
        {Nocentini2023Learning}
\bibfield{author}{\bibinfo{person}{Claudio~Ferrari Nocentini, Federico} {and} \bibinfo{person}{Stefano Berretti.}} \bibinfo{year}{2023}\natexlab{}.
\newblock \showarticletitle{Learning Landmarks Motion from Speech for Speaker-Agnostic 3D Talking Heads Generation.}
\newblock \bibinfo{journal}{\emph{International Conference on Image Analysis and Processing.}} (\bibinfo{year}{2023}).
\newblock


\bibitem[Nocentini et~al\mbox{.}(2024a)]%
        {nocentini2024scantalk}
\bibfield{author}{\bibinfo{person}{Federico Nocentini}, \bibinfo{person}{Thomas Besnier}, \bibinfo{person}{Claudio Ferrari}, \bibinfo{person}{Sylvain Arguillere}, \bibinfo{person}{Stefano Berretti}, {and} \bibinfo{person}{Mohamed Daoudi}.} \bibinfo{year}{2024}\natexlab{a}.
\newblock \showarticletitle{ScanTalk: 3D Talking Heads from Unregistered Scans.}
\newblock \bibinfo{journal}{\emph{arXiv preprint arXiv:2403.10942}} (\bibinfo{year}{2024}).
\newblock


\bibitem[Nocentini et~al\mbox{.}(2024b)]%
        {Nocentini2024EmoVOCA}
\bibfield{author}{\bibinfo{person}{Federico Nocentini}, \bibinfo{person}{Claudio Ferrari}, {and} \bibinfo{person}{Stefano Berretti}.} \bibinfo{year}{2024}\natexlab{b}.
\newblock \showarticletitle{EmoVOCA: Speech-Driven Emotional 3D Talking Heads.}
\newblock \bibinfo{journal}{\emph{arXiv preprint arXiv:2403.12886}} (\bibinfo{year}{2024}).
\newblock


\bibitem[Otberdout et~al\mbox{.}(2020)]%
        {Otberdout2020Dynamic}
\bibfield{author}{\bibinfo{person}{Naima Otberdout}, \bibinfo{person}{Mohamed Daoudi}, \bibinfo{person}{Anis Kacem}, \bibinfo{person}{Lahoucine Ballihi}, {and} \bibinfo{person}{Stefano Berretti}.} \bibinfo{year}{2020}\natexlab{}.
\newblock \showarticletitle{Dynamic facial expression generation on hilbert hypersphere with conditional wasserstein generative adversarial nets.}
\newblock \bibinfo{journal}{\emph{IEEE Transactions on Pattern Analysis and Machine Intelligence 44.2}} (\bibinfo{year}{2020}), \bibinfo{pages}{848--863}.
\newblock


\bibitem[Otberdout et~al\mbox{.}(2022)]%
        {Otberdout2022Sparse}
\bibfield{author}{\bibinfo{person}{Naima Otberdout}, \bibinfo{person}{Claudio Ferrari}, \bibinfo{person}{Mohamed Daoudi}, \bibinfo{person}{Stefano Berretti}, {and} \bibinfo{person}{Alberto Del~Bimbo}.} \bibinfo{year}{2022}\natexlab{}.
\newblock \showarticletitle{Sparse to dense dynamic 3d facial expression generation.}
\newblock \bibinfo{journal}{\emph{Proceedings of the IEEE/CVF conference on computer vision and pattern recognition}} (\bibinfo{year}{2022}).
\newblock


\bibitem[Otberdout et~al\mbox{.}(2023)]%
        {Otberdout2023Generating}
\bibfield{author}{\bibinfo{person}{Naima Otberdout}, \bibinfo{person}{Claudio Ferrari}, \bibinfo{person}{Mohamed Daoudi}, \bibinfo{person}{Stefano Berretti}, {and} \bibinfo{person}{Alberto Del~Bimbo}.} \bibinfo{year}{2023}\natexlab{}.
\newblock \showarticletitle{Generating Multiple 4D Expression Transitions by Learning Face Landmark Trajectories.}
\newblock \bibinfo{journal}{\emph{IEEE Transactions on Affective Computing}} (\bibinfo{year}{2023}).
\newblock


\bibitem[Papantoniou et~al\mbox{.}(2022)]%
        {Papantoniou2022Neural}
\bibfield{author}{\bibinfo{person}{Foivos~Paraperas Papantoniou}, \bibinfo{person}{Panagiotis P.~Filntisis}, \bibinfo{person}{Petros Maragos}, {and} \bibinfo{person}{Anastasios Roussos}.} \bibinfo{year}{2022}\natexlab{}.
\newblock \showarticletitle{Neural Emotion Director: Speech-preserving semantic control of facial expressions in" in-the-wild.}
\newblock \bibinfo{journal}{\emph{In Proceedings of the IEEE/CVF Conference on Computer Vision and Pattern Recognition}} (\bibinfo{year}{2022}), \bibinfo{pages}{18781--18790}.
\newblock


\bibitem[Peng et~al\mbox{.}(2023a)]%
        {Peng2023Selftalk}
\bibfield{author}{\bibinfo{person}{Ziqiao Peng}, \bibinfo{person}{Yihao Luo}, \bibinfo{person}{Yue Shi}, \bibinfo{person}{Hao Xu}, \bibinfo{person}{Xiangyu Zhu}, \bibinfo{person}{Hongyan Liu}, \bibinfo{person}{Jun He}, {and} \bibinfo{person}{Zhaoxin Fan}.} \bibinfo{year}{2023}\natexlab{a}.
\newblock \showarticletitle{Selftalk: A self-supervised commutative training diagram to comprehend 3d talking faces.}
\newblock \bibinfo{journal}{\emph{In Proceedings of the 31st ACM International Conference on Multimedia}} (\bibinfo{year}{2023}), \bibinfo{pages}{5292--5301}.
\newblock


\bibitem[Peng et~al\mbox{.}(2023b)]%
        {Peng2023Emotalk}
\bibfield{author}{\bibinfo{person}{Ziqiao Peng}, \bibinfo{person}{Haoyu Wu}, \bibinfo{person}{Zhenbo Song}, \bibinfo{person}{Hao Xu}, \bibinfo{person}{Xiangyu Zhu}, \bibinfo{person}{Jun He}, \bibinfo{person}{Hongyan Liu}, {and} \bibinfo{person}{Zhaoxin Fan}.} \bibinfo{year}{2023}\natexlab{b}.
\newblock \showarticletitle{Emotalk: Speech-driven emotional disentanglement for 3d face animation.}
\newblock \bibinfo{journal}{\emph{Proceedings of the IEEE/CVF International Conference on Computer Vision}} (\bibinfo{year}{2023}).
\newblock


\bibitem[Rai et~al\mbox{.}(2024)]%
        {Rai2024Towards}
\bibfield{author}{\bibinfo{person}{Aashish Rai}, \bibinfo{person}{Hiresh Gupta}, \bibinfo{person}{Ayush Pandey}, \bibinfo{person}{Francisco Vicente~Carrasco}, \bibinfo{person}{Shingo~Jason Takagi}, \bibinfo{person}{Amaury Aubel}, \bibinfo{person}{Daeil Kim}, \bibinfo{person}{Aayush Prakash}, {and} \bibinfo{person}{Fernando de~la Torre}.} \bibinfo{year}{2024}\natexlab{}.
\newblock \showarticletitle{Towards realistic generative 3d face models.}
\newblock \bibinfo{journal}{\emph{Proceedings of the IEEE/CVF Winter Conference on Applications of Computer Vision}} (\bibinfo{year}{2024}).
\newblock


\bibitem[Ranjan(2018)]%
        {Ranjan2018Generating}
\bibfield{author}{\bibinfo{person}{et~al. Ranjan, Anurag}.} \bibinfo{year}{2018}\natexlab{}.
\newblock \showarticletitle{Generating 3D faces using convolutional mesh autoencoders.}
\newblock \bibinfo{journal}{\emph{Proceedings of the European conference on computer vision (ECCV).}} (\bibinfo{year}{2018}).
\newblock


\bibitem[Richard et~al\mbox{.}(2021)]%
        {Richard2021Meshtalk}
\bibfield{author}{\bibinfo{person}{Alexander Richard}, \bibinfo{person}{Michael Zollhoefer}, \bibinfo{person}{Yandong Wen}, \bibinfo{person}{Fernando de~la Torre}, {and} \bibinfo{person}{Yaser Sheikh}.} \bibinfo{year}{2021}\natexlab{}.
\newblock \showarticletitle{Meshtalk: 3d face animation from speech using cross-modality disentanglement.}
\newblock \bibinfo{journal}{\emph{Proceedings of the IEEE/CVF International Conference on Computer Vision}} (\bibinfo{year}{2021}).
\newblock


\bibitem[Saunders and Namboodiri(2023)]%
        {saunders2023read}
\bibfield{author}{\bibinfo{person}{Jack Saunders} {and} \bibinfo{person}{Vinay Namboodiri}.} \bibinfo{year}{2023}\natexlab{}.
\newblock \showarticletitle{Read avatars: Realistic emotion-controllable audio driven avatars.}
\newblock \bibinfo{journal}{\emph{arXiv preprint arXiv:2303.00744}} (\bibinfo{year}{2023}).
\newblock


\bibitem[Schuller et~al\mbox{.}(2003)]%
        {Schuller2003Hidden}
\bibfield{author}{\bibinfo{person}{Björn Schuller}, \bibinfo{person}{Gerhard Rigoll}, {and} \bibinfo{person}{Manfred Lang}.} \bibinfo{year}{2003}\natexlab{}.
\newblock \showarticletitle{Hidden Markov model-based speech emotion recognition.}
\newblock \bibinfo{journal}{\emph{In 2003 IEEE International Conference on Acoustics, Speech, and Signal Processing, 2003. Proceedings. (ICASSP'03), vol. 2, pp. II-1, Ieee}} (\bibinfo{year}{2003}).
\newblock


\bibitem[Schuller(2018)]%
        {Schuller2018Speech}
\bibfield{author}{\bibinfo{person}{Björn~W Schuller}.} \bibinfo{year}{2018}\natexlab{}.
\newblock \showarticletitle{Speech emotion recognition: Two decades in a nutshell, benchmarks, and ongoing trends.}
\newblock \bibinfo{journal}{\emph{Communications of the ACM 61}} \bibinfo{number}{5} (\bibinfo{year}{2018}), \bibinfo{pages}{90--99}.
\newblock


\bibitem[Shazeer(2019)]%
        {Shazeer2019Fast}
\bibfield{author}{\bibinfo{person}{Noam Shazeer}.} \bibinfo{year}{2019}\natexlab{}.
\newblock \showarticletitle{Fast transformer decoding: One write-head is all you need.}
\newblock \bibinfo{journal}{\emph{arXiv preprint arXiv:1911.02150}} (\bibinfo{year}{2019}).
\newblock


\bibitem[Shen et~al\mbox{.}(2024)]%
        {Shen2024DEITalk}
\bibfield{author}{\bibinfo{person}{Kang Shen}, \bibinfo{person}{Haifeng Xia}, \bibinfo{person}{Guangxing Geng}, \bibinfo{person}{GuangYue Geng}, \bibinfo{person}{Siyu Xia}, {and} \bibinfo{person}{Zhengming Ding}.} \bibinfo{year}{2024}\natexlab{}.
\newblock \showarticletitle{DEITalk: Speech-Driven 3D Facial Animation with Dynamic Emotional Intensity Modeling.}
\newblock \bibinfo{journal}{\emph{In ACM Multimedia}} (\bibinfo{year}{2024}).
\newblock


\bibitem[Shen et~al\mbox{.}(2023)]%
        {Shen2023Difftalk}
\bibfield{author}{\bibinfo{person}{Shuai Shen}, \bibinfo{person}{Wenliang Zhao}, \bibinfo{person}{Zibin Meng}, \bibinfo{person}{Wanhua Li}, \bibinfo{person}{Zheng Zhu}, \bibinfo{person}{Jie Zhou}, {and} \bibinfo{person}{Jiwen Lu}.} \bibinfo{year}{2023}\natexlab{}.
\newblock \showarticletitle{Difftalk: Crafting diffusion models for generalized audio-driven portraits animation.}
\newblock \bibinfo{journal}{\emph{Proceedings of the IEEE/CVF Conference on Computer Vision and Pattern Recognition}} (\bibinfo{year}{2023}).
\newblock


\bibitem[Song et~al\mbox{.}(2022)]%
        {Song2022Talking}
\bibfield{author}{\bibinfo{person}{Hyoung-Kyu Song}, \bibinfo{person}{Sang Hoon~Woo}, \bibinfo{person}{Junhyeok Lee}, \bibinfo{person}{Seungmin Yang}, \bibinfo{person}{Hyunjae Cho}, \bibinfo{person}{Youseong Lee}, \bibinfo{person}{Dongho Choi}, {and} \bibinfo{person}{Kang-wook Kim}.} \bibinfo{year}{2022}\natexlab{}.
\newblock \showarticletitle{Talking face generation with multilingual tts.}
\newblock \bibinfo{journal}{\emph{Proceedings of the IEEE/CVF Conference on Computer Vision and Pattern Recognition}} (\bibinfo{year}{2022}).
\newblock


\bibitem[Stan et~al\mbox{.}(2023)]%
        {Stan2023Facediffuser}
\bibfield{author}{\bibinfo{person}{Stefan Stan}, \bibinfo{person}{Kazi Injamamul~Haque}, {and} \bibinfo{person}{Zerrin Yumak}.} \bibinfo{year}{2023}\natexlab{}.
\newblock \showarticletitle{Facediffuser: Speech-driven 3d facial animation synthesis using diffusion.}
\newblock \bibinfo{journal}{\emph{Proceedings of the 16th ACM SIGGRAPH Conference on Motion, Interaction, and Games}} (\bibinfo{year}{2023}).
\newblock


\bibitem[Stypułkowski et~al\mbox{.}(2024)]%
        {Stypułkowski2024Diffused}
\bibfield{author}{\bibinfo{person}{Michał Stypułkowski}, \bibinfo{person}{Konstantinos Vougioukas}, \bibinfo{person}{Sen He}, \bibinfo{person}{Maciej Zięba}, \bibinfo{person}{Stavros Petridis}, {and} \bibinfo{person}{Maja Pantic}.} \bibinfo{year}{2024}\natexlab{}.
\newblock \showarticletitle{Diffused heads: Diffusion models beat gans on talking-face generation.}
\newblock \bibinfo{journal}{\emph{Proceedings of the IEEE/CVF Winter Conference on Applications of Computer Vision}} (\bibinfo{year}{2024}).
\newblock


\bibitem[Su et~al\mbox{.}(2023)]%
        {Su2023DualTalker}
\bibfield{author}{\bibinfo{person}{Guinan Su}, \bibinfo{person}{Yanwu Yang}, {and} \bibinfo{person}{Zhifeng Li}.} \bibinfo{year}{2023}\natexlab{}.
\newblock \showarticletitle{DualTalker: A Cross-Modal Dual Learning Approach for Speech-Driven 3D Facial Animation.}
\newblock \bibinfo{journal}{\emph{arXiv preprint arXiv:2311.04766}} (\bibinfo{year}{2023}).
\newblock


\bibitem[Su et~al\mbox{.}(2024)]%
        {Su2024Roformer}
\bibfield{author}{\bibinfo{person}{Jianlin Su}, \bibinfo{person}{Murtadha Ahmed}, \bibinfo{person}{Yu Lu}, \bibinfo{person}{Bo~Wen Pan, Shengfeng}, {and} \bibinfo{person}{Yunfeng Liu}.} \bibinfo{year}{2024}\natexlab{}.
\newblock \showarticletitle{Roformer: Enhanced transformer with rotary position embedding.}
\newblock \bibinfo{journal}{\emph{Neurocomputing}} \bibinfo{number}{568} (\bibinfo{year}{2024}), \bibinfo{pages}{127063}.
\newblock


\bibitem[Sun et~al\mbox{.}(2023)]%
        {Sun2023Vividtalk}
\bibfield{author}{\bibinfo{person}{Xusen Sun}, \bibinfo{person}{Longhao Zhang}, \bibinfo{person}{Hao Zhu}, \bibinfo{person}{Peng Zhang}, \bibinfo{person}{Bang Zhang}, \bibinfo{person}{Xinya Ji}, \bibinfo{person}{Kangneng Zhou}, \bibinfo{person}{Daiheng Gao}, \bibinfo{person}{Liefeng Bo}, {and} \bibinfo{person}{Xun Cao}.} \bibinfo{year}{2023}\natexlab{}.
\newblock \showarticletitle{Vividtalk: One-shot audio-driven talking head generation based on 3d hybrid prior.}
\newblock \bibinfo{journal}{\emph{arXiv - CS - Computer Vision and Pattern Recognition}} (\bibinfo{year}{2023}).
\newblock


\bibitem[Sun et~al\mbox{.}(2022)]%
        {Sun2022Masked}
\bibfield{author}{\bibinfo{person}{Yasheng Sun}, \bibinfo{person}{Hang Zhou}, \bibinfo{person}{Kaisiyuan Wang}, \bibinfo{person}{Qianyi Wu}, \bibinfo{person}{Zhibin Hong}, \bibinfo{person}{Jingtuo Liu}, \bibinfo{person}{Errui Ding}, \bibinfo{person}{Jingdong Wang}, \bibinfo{person}{Ziwei Liu}, {and} \bibinfo{person}{Koike Hideki}.} \bibinfo{year}{2022}\natexlab{}.
\newblock \showarticletitle{Masked lip-sync prediction by audio-visual contextual exploitation in transformers.}
\newblock \bibinfo{journal}{\emph{SIGGRAPH Asia 2022 Conference Papers}} (\bibinfo{year}{2022}).
\newblock


\bibitem[Sun et~al\mbox{.}(2024)]%
        {Sun2024Diffposetalk}
\bibfield{author}{\bibinfo{person}{Zhiyao Sun}, \bibinfo{person}{Tian Lv}, \bibinfo{person}{Sheng Ye}, \bibinfo{person}{Matthieu Lin}, \bibinfo{person}{Jenny Sheng}, \bibinfo{person}{Yu-Hui Wen},{and} \bibinfo{person}{Yong-jin Liu}.} \bibinfo{year}{2024}\natexlab{}.
\newblock \showarticletitle{Diffposetalk: Speech-driven stylistic 3d facial animation and head pose generation via diffusion models.}
\newblock \bibinfo{journal}{\emph{CM Transactions on Graphics (TOG)}} (\bibinfo{year}{2024}).
\newblock


\bibitem[Sung-Bin et~al\mbox{.}(2024)]%
        {Sung-Bin2024LaughTalk}
\bibfield{author}{\bibinfo{person}{Kim Sung-Bin}, \bibinfo{person}{Lee Hyun}, \bibinfo{person}{Da Hye~Hong}, \bibinfo{person}{Suekyeong Nam}, \bibinfo{person}{Janghoon Ju}, {and} \bibinfo{person}{Tae-Hyun Oh}.} \bibinfo{year}{2024}\natexlab{}.
\newblock \showarticletitle{LaughTalk: Expressive 3D Talking Head Generation with Laughter.}
\newblock \bibinfo{journal}{\emph{Proceedings of the IEEE/CVF Winter Conference on Applications of Computer Vision}} (\bibinfo{year}{2024}).
\newblock


\bibitem[Tan et~al\mbox{.}(2024d)]%
        {Tan2024Say}
\bibfield{author}{\bibinfo{person}{Shuai Tan}, \bibinfo{person}{Bin Ji}, \bibinfo{person}{Yu Ding}, {and} \bibinfo{person}{Ye Pan}.} \bibinfo{year}{2024}\natexlab{d}.
\newblock \showarticletitle{Say anything with any style."}.
\newblock \bibinfo{journal}{\emph{Proceedings of the AAAI Conference on Artificial Intelligence}} \bibinfo{volume}{38}, \bibinfo{number}{5} (\bibinfo{year}{2024}).
\newblock


\bibitem[Tan et~al\mbox{.}(2024a)]%
        {tan2024flowvqtalker}
\bibfield{author}{\bibinfo{person}{Shuai Tan}, \bibinfo{person}{Bin Ji}, {and} \bibinfo{person}{Ye Pan}.} \bibinfo{year}{2024}\natexlab{a}.
\newblock \showarticletitle{FlowVQTalker: High-Quality Emotional Talking Face Generation through Normalizing Flow and Quantization.}
\newblock \bibinfo{journal}{\emph{Proceedings of the IEEE/CVF Conference on Computer Vision and Pattern Recognition}} (\bibinfo{year}{2024}).
\newblock


\bibitem[Tan et~al\mbox{.}(2024b)]%
        {Tan2024Style2talker}
\bibfield{author}{\bibinfo{person}{Shuai Tan}, \bibinfo{person}{Bin Ji}, {and} \bibinfo{person}{Pan Ye}.} \bibinfo{year}{2024}\natexlab{b}.
\newblock \showarticletitle{Style2talker: High-resolution talking head generation with emotion style and art style.}
\newblock \bibinfo{journal}{\emph{Proceedings of the AAAI Conference on Artificial Intelligence}} \bibinfo{volume}{38}, \bibinfo{number}{5} (\bibinfo{year}{2024}).
\newblock


\bibitem[Tan et~al\mbox{.}(2024c)]%
        {Tan2024EDTalk}
\bibfield{author}{\bibinfo{person}{Shuai Tan}, \bibinfo{person}{Bin~Ji Ji}, \bibinfo{person}{Mengxiao Bi}, {and} \bibinfo{person}{Ye Pan}.} \bibinfo{year}{2024}\natexlab{c}.
\newblock \showarticletitle{EDTalk: Efficient Disentanglement for Emotional Talking Head Synthesis.}
\newblock \bibinfo{journal}{\emph{arXiv preprint arXiv:2404.01647}} (\bibinfo{year}{2024}).
\newblock


\bibitem[Tang et~al\mbox{.}(2022)]%
        {Tang2022Memories}
\bibfield{author}{\bibinfo{person}{Anni Tang}, \bibinfo{person}{Tianyu He}, \bibinfo{person}{Xu Tan}, \bibinfo{person}{Jun Ling}, \bibinfo{person}{Runnan Li}, \bibinfo{person}{Sheng Zhao}, \bibinfo{person}{Li Song}, {and} \bibinfo{person}{Jiang Bian}.} \bibinfo{year}{2022}\natexlab{}.
\newblock \showarticletitle{Memories are one-to-many mapping alleviators in talking face generation.}
\newblock \bibinfo{journal}{\emph{arXiv preprint arXiv:2212.05005}} (\bibinfo{year}{2022}).
\newblock


\bibitem[Tian et~al\mbox{.}(2024)]%
        {tian2024emo}
\bibfield{author}{\bibinfo{person}{Linrui Tian}, \bibinfo{person}{Qi Wang}, \bibinfo{person}{Bang Zhang}, {and} \bibinfo{person}{Liefeng Bo}.} \bibinfo{year}{2024}\natexlab{}.
\newblock \showarticletitle{EMO: Emote Portrait Alive-Generating Expressive Portrait Videos with Audio2Video Diffusion Model under Weak Conditions.}
\newblock \bibinfo{journal}{\emph{arXiv preprint arXiv:2402.17485}} (\bibinfo{year}{2024}).
\newblock


\bibitem[Wang et~al\mbox{.}(2023)]%
        {Wang2023Seeing}
\bibfield{author}{\bibinfo{person}{Jiadong Wang}, \bibinfo{person}{Xinyuan Qian}, \bibinfo{person}{Malu Zhang}, \bibinfo{person}{Robby T.~Tan}, {and} \bibinfo{person}{Haizhou Li}.} \bibinfo{year}{2023}\natexlab{}.
\newblock \showarticletitle{Seeing what you said: Talking face generation guided by a lip-reading expert.}
\newblock \bibinfo{journal}{\emph{Proceedings of the IEEE/CVF Conference on Computer Vision and Pattern Recognition}} (\bibinfo{year}{2023}).
\newblock


\bibitem[Wei et~al\mbox{.}(2024)]%
        {Wei2024Aniportrait}
\bibfield{author}{\bibinfo{person}{Huawei Wei}, \bibinfo{person}{Zejun Yang}, {and} \bibinfo{person}{Zhisheng Wang}.} \bibinfo{year}{2024}\natexlab{}.
\newblock \showarticletitle{Aniportrait: Audio-driven synthesis of photorealistic portrait animation.}
\newblock \bibinfo{journal}{\emph{arXiv preprint arXiv:2403.17694}} (\bibinfo{year}{2024}).
\newblock


\bibitem[Xing et~al\mbox{.}(2023)]%
        {xing2023codetalker}
\bibfield{author}{\bibinfo{person}{Jinbo Xing}, \bibinfo{person}{Menghan Xia}, \bibinfo{person}{Yuechen Zhang}, \bibinfo{person}{Xiaodong Cun}, \bibinfo{person}{Jue Wang}, {and} \bibinfo{person}{Tien-Tsin Wong}.} \bibinfo{year}{2023}\natexlab{}.
\newblock \showarticletitle{Codetalker: Speech-driven 3d facial animation with discrete motion prior.}
\newblock \bibinfo{journal}{\emph{Proceedings of the IEEE/CVF Conference on Computer Vision and Pattern Recognition}} (\bibinfo{year}{2023}).
\newblock


\bibitem[Xu et~al\mbox{.}(2024d)]%
        {xu2024facechain-imagineid}
\bibfield{author}{\bibinfo{person}{Chao Xu}, \bibinfo{person}{Yang Liu}, \bibinfo{person}{Jiazheng Xing}, \bibinfo{person}{Weida Wang}, \bibinfo{person}{Mingze Sun}, \bibinfo{person}{Jun Dan}, {and} \bibinfo{person}{Tianxin Huang}.} \bibinfo{year}{2024}\natexlab{d}.
\newblock \showarticletitle{FaceChain-ImagineID: Freely Crafting High-Fidelity Diverse Talking Faces from Disentangled Audio.}
\newblock \bibinfo{journal}{\emph{Proceedings of the IEEE/CVF Conference on Computer Vision and Pattern Recognition}} (\bibinfo{year}{2024}).
\newblock


\bibitem[Xu et~al\mbox{.}(2024b)]%
        {Xu2024Hallo}
\bibfield{author}{\bibinfo{person}{Mingwang Xu}, \bibinfo{person}{Hui Li}, \bibinfo{person}{Qingkun Su}, \bibinfo{person}{Hanlin Shang}, \bibinfo{person}{Liwei Zhang}, \bibinfo{person}{Ce Liu}, \bibinfo{person}{Van Gool~Luc Wang, Jingdong}, \bibinfo{person}{Yao Yao}, {and} \bibinfo{person}{Siyu Zhu}.} \bibinfo{year}{2024}\natexlab{b}.
\newblock \showarticletitle{Hallo: Hierarchical Audio-Driven Visual Synthesis for Portrait Image Animation.}
\newblock \bibinfo{journal}{\emph{arXiv preprint arXiv:2406.08801}} (\bibinfo{year}{2024}).
\newblock


\bibitem[Xu et~al\mbox{.}(2024a)]%
        {Xu2024Vase-1}
\bibfield{author}{\bibinfo{person}{Sicheng Xu}, \bibinfo{person}{Guojun Chen}, \bibinfo{person}{Yu-Xiao Guo}, \bibinfo{person}{Jiaolong Yang}, \bibinfo{person}{Chong Li}, \bibinfo{person}{Zhenyu Zang}, \bibinfo{person}{Yizhong Zhang}, \bibinfo{person}{Xin Tong}, {and} \bibinfo{person}{Baining Guo}.} \bibinfo{year}{2024}\natexlab{a}.
\newblock \showarticletitle{Vasa-1: Lifelike audio-driven talking faces generated in real-time.}
\newblock \bibinfo{journal}{\emph{arXiv preprint arXiv:2404.10667}} (\bibinfo{year}{2024}).
\newblock


\bibitem[Xu et~al\mbox{.}(2024c)]%
        {Xu2024Mambatalk}
\bibfield{author}{\bibinfo{person}{Zunnan Xu}, \bibinfo{person}{Yukang Lin}, \bibinfo{person}{Haonan Han}, \bibinfo{person}{Sicheng Yang}, \bibinfo{person}{Ronghui Li}, \bibinfo{person}{Yachao Zhang}, {and} \bibinfo{person}{Xiu Li}.} \bibinfo{year}{2024}\natexlab{c}.
\newblock \showarticletitle{Mambatalk: Efficient holistic gesture synthesis with selective state space models.}
\newblock \bibinfo{journal}{\emph{arXiv preprint arXiv:2403.09471}} (\bibinfo{year}{2024}).
\newblock


\bibitem[Yang et~al\mbox{.}(2025)]%
        {Yang2025StyleSpeaker}
\bibfield{author}{\bibinfo{person}{An Yang}, \bibinfo{person}{Chenyu Liu}, {and} \bibinfo{person}{Jun Du}.} \bibinfo{year}{2025}\natexlab{}.
\newblock \showarticletitle{StyleSpeaker: Audio-Enhanced Fine-Grained Style Modeling for Speech-Driven 3D Facial Animation.}
\newblock \bibinfo{journal}{\emph{arXiv preprint arXiv:2503.09852}} (\bibinfo{year}{2025}).
\newblock


\bibitem[Ye et~al\mbox{.}(2023)]%
        {Ye2023Geneface++}
\bibfield{author}{\bibinfo{person}{Zhenhui Ye}, \bibinfo{person}{Jinzheng He}, \bibinfo{person}{Ziyue Jiang}, \bibinfo{person}{Rongjie Huang}, \bibinfo{person}{Jiawei Huang}, \bibinfo{person}{Ren Yi Yin Xiang Ma~Zejun Liu, Jinglin}, {and} \bibinfo{person}{Zhou Zhao}.} \bibinfo{year}{2023}\natexlab{}.
\newblock \showarticletitle{Geneface++: Generalized and stable real-time audio-driven 3d talking face generation.}
\newblock \bibinfo{journal}{\emph{arXiv preprint arXiv:2305.00787}} (\bibinfo{year}{2023}).
\newblock


\bibitem[Ye et~al\mbox{.}(2024)]%
        {Ye2024Real3d-portrait}
\bibfield{author}{\bibinfo{person}{Zhenhui Ye}, \bibinfo{person}{Tianyun Zhong}, \bibinfo{person}{Yi Ren}, \bibinfo{person}{Jiaqi Yang}, \bibinfo{person}{Weichuang Li}, \bibinfo{person}{Jiawei Huang}, \bibinfo{person}{Ziyue Jiang}, {and} \bibinfo{person}{et al.}} \bibinfo{year}{2024}\natexlab{}.
\newblock \showarticletitle{Real3d-portrait: One-shot realistic 3d talking portrait synthesis.}
\newblock \bibinfo{journal}{\emph{arXiv preprint arXiv:2401.08503}} (\bibinfo{year}{2024}).
\newblock


\bibitem[Zhong et~al\mbox{.}(2023)]%
        {Zhong2023Identity-preserving}
\bibfield{author}{\bibinfo{person}{Weizhi Zhong}, \bibinfo{person}{Chaowei Fang}, \bibinfo{person}{Yinqi Cai}, \bibinfo{person}{Pengxu Wei}, \bibinfo{person}{Gangming Zhao}, \bibinfo{person}{Liang Lin}, {and} \bibinfo{person}{Guanbin Li}.} \bibinfo{year}{2023}\natexlab{}.
\newblock \showarticletitle{Identity-preserving talking face generation with landmark and appearance priors.}
\newblock \bibinfo{journal}{\emph{Proceedings of the IEEE/CVF Conference on Computer Vision and Pattern Recognition}} (\bibinfo{year}{2023}).
\newblock


\bibitem[Zou et~al\mbox{.}(2023)]%
        {Zou20234D}
\bibfield{author}{\bibinfo{person}{Kaifeng Zou}, \bibinfo{person}{Sylvain Faisan}, \bibinfo{person}{Boyang Yu}, \bibinfo{person}{Sébastien Valette}, {and} \bibinfo{person}{Hyewon Seo}.} \bibinfo{year}{2023}\natexlab{}.
\newblock \showarticletitle{4d facial expression diffusion model.}
\newblock \bibinfo{journal}{\emph{ACM Transactions on Multimedia Computing, Communications, and Applications}} (\bibinfo{year}{2023}).
\newblock


\end{thebibliography}
\end{document}